\documentclass[journal,twoside,web]{ieeecolor}
\usepackage{jsen}
\usepackage{cite}
\usepackage{amsmath,amssymb,amsfonts}
\usepackage{algorithmic}
\usepackage{graphicx}
\usepackage{textcomp}
\usepackage{wrapfig}
\usepackage{gensymb}
\usepackage{siunitx}
\usepackage{amsmath}
\usepackage{multirow}
\usepackage{makecell}
\usepackage{subfigure}
\usepackage{nicefrac, xfrac}
\usepackage[dvipsnames]{xcolor}
\usepackage{ulem}

\newcolumntype{P}[1]{>{\centering\arraybackslash}p{#1}}
\newcolumntype{M}[1]{>{\centering\arraybackslash}m{#1}}
\def\BibTeX{{\rm B\kern-.05em{\sc i\kern-.025em b}\kern-.08em
    T\kern-.1667em\lower.7ex\hbox{E}\kern-.125emX}}
\definecolor{abstractbg}{rgb}{0.89804,0.94510,0.83137}
\begin{document}
\title{Design and Fabrication of a Fiber Bragg Grating Shape Sensor for Shape Reconstruction of a Continuum Manipulator} %Continuum Manipulator} 
\author{Golchehr Amirkhani, Anna Goodridge, Mojtaba Esfandiari, Henry Phalen, \IEEEmembership{Graduate Student Member, IEEE}, Justin H. Ma., \IEEEmembership{Graduate Student Member, IEEE}, Iulian Iordachita, \IEEEmembership{Senior Member, IEEE}, and Mehran Armand, \IEEEmembership{Member, IEEE}
\thanks{%This paragraph of the first footnote will contain the date on 
%which you submitted your paper for review. It %will also contain support 
%information, including sponsor and financial support acknowledgment. For 
%example, 
This work was supported in part by the National 
Institutes of Health (NIH) under Grant R01 EB016703 and R01 AR080315.}
\thanks{Golchehr Amirkhani, Mojtaba Esfandiari, Henry Phalen, Justin H. Ma, and Iulian Iordachita are with the Department of Mechanical Engineering, Johns Hopkins University, Baltimore, MD 21218 USA, and also with the Laboratory for Computational Sensing and Robotics, Johns Hopkins University, Baltimore, MD, 21218 USA (e-mail: gamirkh1@jhu.edu; mesfand2@jhu.edu; henry.phalen@jhu.edu; jma60@jhu.edu; iordachita@jhu.edu).
}
\thanks{Anna Goodridge is with the Laboratory for Computational Sensing and Robotics, Johns Hopkins University, Baltimore, MD, 21218 USA (e-mail: anna.goodridge@jhu.edu).
}
\thanks{Mehran Armand is with the Department of orthopaedic Surgery, the Department of Mechanical Engineering,  the Department of Computer Science, Johns, and also with the Laboratory for Computational Sensing and Robotics, Johns Hopkins University, Baltimore, MD, 21218 USA  (e-mail: marmand2@jhu.edu).
}
}

\IEEEtitleabstractindextext{%
\fcolorbox{abstractbg}{abstractbg}{%
\begin{minipage}{\textwidth}%
\begin{wrapfigure}[12]{r}{3in}%
\includegraphics[width=3in]{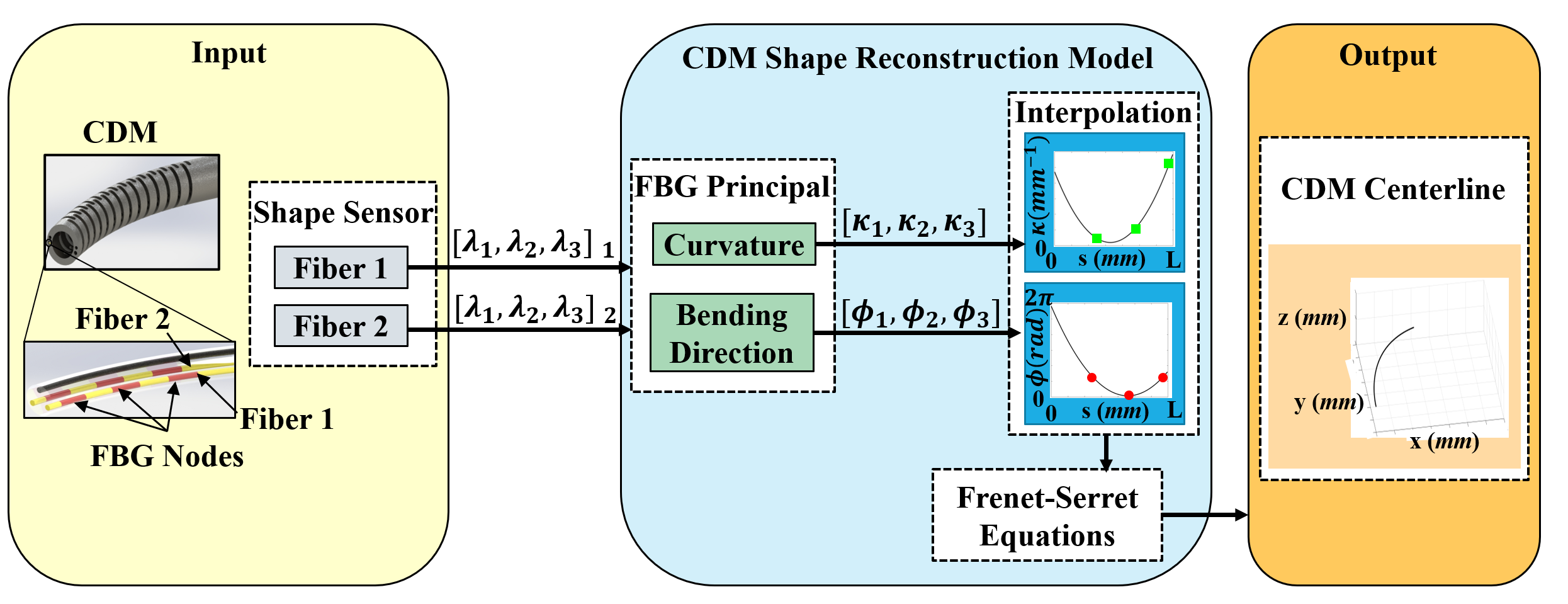}%
\end{wrapfigure}%
\begin{abstract}
Continuum dexterous manipulators (CDMs) are suitable
for performing tasks in a constrained environment due to their high
dexterity and maneuverability. Despite the inherent advantages of
CDMs in minimally invasive surgery, real-time control of CDMs' shape during non-constant curvature bending is still challenging. This study presents a novel approach for the design and fabrication of a large deflection fiber Bragg grating (FBG) shape sensor embedded within the lumens inside the walls of a CDM with a large instrument channel. The shape sensor consisted of two fibers, each with three FBG nodes. A shape-sensing model was introduced to reconstruct the centerline of the CDM based on FBG wavelengths. Different experiments, including shape sensor tests and CDM shape reconstruction tests, were conducted to assess the overall accuracy of the shape sensing. The FBG sensor evaluation results revealed the linear curvature-wavelength relationship with the large curvature detection of
0.045 mm at a 90$^{\circ}$ bending angle and a sensitivity of up to 5.50 \nicefrac{nm}{mm} in each bending direction. The CDM’s shape reconstruction experiments in a free environment demonstrated the shape tracking accuracy of 0.216\textpm0.126 mm for positive/negative deflections. Also, the CDM shape reconstruction error for three cases of bending with obstacles were observed to be 0.436\textpm0.370 mm for the proximal case, 0.485\textpm0.418 mm for the middle case, and 0.312\textpm0.261 mm for the distal case. This study indicates the adequate performance of the FBG sensor and the effectiveness of
the model for tracking the shape of the large-deflection CDM with nonconstant-curvature bending for minimally-invasive orthopaedic  applications.
\end{abstract}

\end{minipage}}}

\maketitle

\section{Introduction}
\label{sec:introduction}
\IEEEPARstart{I}{n} recent years, continuum manipulators have shown great potential to improve steerability and dexterity in minimally invasive robotic surgery (MIRS) procedures. The compliant structure of CDMs offer flexibility for reaching the targeted site of treatment in a constrained environment as their bodies can conform to their surroundings \cite{burgner2015continuum,simaan2004high,yip2014model,vitiello2012emerging,amirkhani2020extended}. Despite many advantages of CDMs for medical applications, accurate estimation of its position and shape is a challenging task, especially when operating within unknown/unmodeled environments. 

Approaches for the shape reconstruction of CDMs in surgical scenarios may include model-based, image-based, sensor-based and/or their combination. The model-based approach employs the mechanics and kinematics of CDMs for shape reconstruction \cite{gao2016mechanical,rucker2011statics,xu2010analytic,segreti2012cable,murphy2014design}. Due to the compliance of CDMs, these models may be complex and require a long computational time, making them challenging to use in real-time applications \cite{chirikjian2015conformational,gao2016mechanical}. The accuracy of the model-dependent CDM shape sensing relies on the system parameter identification and assumptions when dealing with unknown disturbances and environmental constraints \cite{webster2010design,moses2015modeling}. 
The image-based approaches, on the other hand, may be more accurate since they do not rely on the mechanics of the CDM \cite{vandini2015vision,hannan2005real}. However, radiation (i.e. fluoroscopy) or lower-quality images (i.e ultrasound imaging) introduce specific challenges for the real-time application of MIRS \cite{lobaton2013continuous,camarillo2008vision}.

The literature reports a number of shape-sensing approaches for real-time control and shape reconstruction of CDMs, including the use of piezoelectric polymers, electromagnetic (EM), and optical sensors \cite{sadjadi2016simultaneous,hill1997fiber,cianchetti2012sensorization}. EM and optical sensors are more commonly used, because of their smaller size, in real-time localization and the shape tracking of CDMs for surgery \cite{feuerstein2009magneto,ryu2014fbg,liu2015shape}. EM tracking systems, however, are susceptible to errors arising from magnetic field distortions caused by the conductive objects in the work field \cite{franz2014electromagnetic}. 

%Real-time shape reconstruction of continuum manipulators is still an ongoing and active research field in MIRS. Sensor-based approach, which is an emerging technique for shape estimation and closed-loop control, has been recently proposed. Electromagnetic (EM) tracking-based shape sensing technique has been utilized in real-time localization and tracking of continuum robots. EM sensors can track both tip and shape by attaching multiple miniature EM sensors along the continuum robots \cite{sadjadi2016simultaneous}. However, EM tracking systems are susceptible to errors arising from magnetic field distortions caused by conductive objects in the working field \cite{franz2014electromagnetic}.% 

The thin, flexible, lightweight and biocompatible structure of optical shape sensors with the added advantage of electromagnetic immunity from external devices has received increasing attention in the sensor-based shape sensing of the flexible instruments. By measuring the reflected wavelength from each fiber, strain and curvature sensing of FBG sensors enable accurate shape-tracking without relying on the kinematics and mechanics of the robotic systems \cite{taffoni2013optical,khan2020pose,hill1997fiber,abushagur2014advances}. Optical fibers such as FBG with different configurations have been employed for the shape reconstruction of flexible instruments such as catheters and biopsy needles as well as CDMs \cite{park2010real,farvardin2016towards,khan2017force,abayazid20133d}. FBG shape sensors, in particular, are classified into two main configurations: A bundle of single-core fibers; and multicore fibers (MCFs). 
%Park \textit{et. al} \cite{park2010real} embedded optical fibers into a biopsy needle with three grooves. The shape sensor consisted of three optical fibers, each with two FBG nodes. The deflected shape of the needle has been detected using the beam theory for curvature calculation from FBG nodes. 
Roesthuis \textit{et. al} \cite{roesthuis2013using} fabricated triplet optical fibers, each with four FBG nodes. Fibers were attached to a three-groove NiTi rod, placed in the backbone of the continuum manipulator. The three-dimensional shape of the manipulator has been reconstructed in terms of the curvature and strain relationship, derived from axial strain measurements of FBG nodes. Liu \textit{et. al} \cite{liu2015shape} developed an FBG shape sensor by attaching a fiber with three FBG nodes to two NiTi rods in a triangular configuration and passing the sensor assembly through the sensor channel of a continuum manipulator. FBG wavelength-curvature calibration was used to find the curvature at discrete locations and then reconstruct the shape of the continuum manipulator. Sefati \textit{et. al} \cite{sefati2016fbg} built an FBG sensor assembly by embedding a fiber and two NiTi rods into a three-lumen polycarbonate tube with circular cross-section. The sensor assembly was passed through the side channel of a CDM. The design was later changed to insert three fibers, each with three FBG nodes, into the three-groove NiTi rod. A data-driven approach was developed to track the tip position of the CDM using FBG wavelength measurements \cite{sefati2017highly,sefati2019fbg}. 
%FBG sensors can also be presented in the configuration of the MCFs, as curvature sensors for the shape reconstruction.
Moore \textit{et. al} \cite{moore2012shape} reconstructed the three-dimensional shape using MCFs by combining the elastic rod theory and differential geometry. Khan \textit{et. al} \cite{khan2019multi} employed several MCFs for sensing the shape of a flexible medical instrument. The shape of the instrument is reconstructed using the curvature and torsion calculated from the FBG wavelength data. Cao \textit{et. al} \cite{cao2022spatial} integrated MCFs into a continuum robot to track the shape of the robot in the free space. 

A major limitation of the previous work on the bundle of single-core fibers is the challenge and fabrication time associated with inserting and keeping fibers in the grooves of the sensor's substrate due to the thin and delicate nature of fibers \cite{park2010real,roesthuis2013using,sefati2017highly}. When the sensing unit is embedded into the channel within the wall of the continuum manipulator, the glue amount applied to the grooves of the sensor's substrate is an important factor. Finding a sufficient amount of glue to not exceed the channel diameter constraint makes the building procedure difficult \cite{roesthuis2013using,sefati2017highly}. Furthermore, attaching the fiber to the outer wall of the substrate is a complex manufacturing process. The sensor configuration becomes trial-dependent, making reproducibility a challenge \cite{liu2015shape}. The enclosed substrate is an appropriate option, especially if the sensor assembly is routed through the channels within the wall of the CDM. However, in the previous study, 
%due to the components of the sensor assembly and the fabrication process, in the previous study, the glue has been applied only at both ends of the sensing unit.
the shape sensor suffered from low sensitivity to small curvature changes, resulting in a low signal-to-noise ratio \cite{sefati2016fbg}. The shape sensor with MCFs, on the other hand, has a smaller cross-section and more accurate FBG alignment \cite{moore2012shape,khan2019multi}. Nevertheless, light coupling into each core is difficult, and MCFs are more expensive than single-core fibers due to the draw tower fabrication procedure \cite{takenaga2010reduction,flockhart2003two}. Moreover, for applications in minimally invasive orthopedic interventions, a bundle of single core fibers is sufficient for detecting the CDM centerline \cite{sefati2016fbg,sefati2017highly,liu2015shape,sefati2019fbg}. Most of the prior work have introduced the FBG-based shape reconstruction algorithm for the case that the shape sensor is directly attached to the flexible medical instrument \cite{park2010real,roesthuis2013using,moore2012shape,khan2019multi,khan2020pose}. The free movement of the sensing unit inside the wall channel of the CDM makes shape tracking difficult. The CDM centerline detection has rarely been developed in previous studies, and the sensing model did not account for the friction effect between the sensor and the CDM channel's wall \cite{sefati2017highly,liu2015shape,sefati2016fbg,cao2022spatial}.

The purpose of this study is to design and build a novel FBG-based shape sensor that is inexpensive, can be easily fabricated and integrated with a CDM undergoing large bending and deflections during orthopaedic procedures. Another contribution includes the formulation of the shape reconstruction model for the CDM, that can significantly lessen the influence of internal twist and friction. The performance of the model is assessed in both free and constrained environments.

%The purpose of this study is to design and fabricate a novel shape sensor using two optical fibers and a NiTi wire embedded in a polycarbonate tube to give an overall diameter of 500 \textit{$\micro$m}. The sensing unit is integrated into the side sensor channel of the CDM such that the shape sensor moves freely when the CDM bends. As a result, the enclosed substrate can act as a shield to prevent adhesive wear. The flexibility of the sensor components guarantees the high-bending up to 90$^\circ$ and the glue coverage through the whole length of the substrate provides high sensitivity for the sensor. The shape reconstruction model is formulated for the CDM, which accounts for the internal twist compensation and the friction compensation in positive and negative deflections, and its performance is assessed in a set of experiments including the free and constraint environments.%

\section{Design}
\label{sec:design}

\subsection{FBG Sensor Design Requirements}
\begin{figure}[!t]
\centering
\includegraphics[width=2.2in]{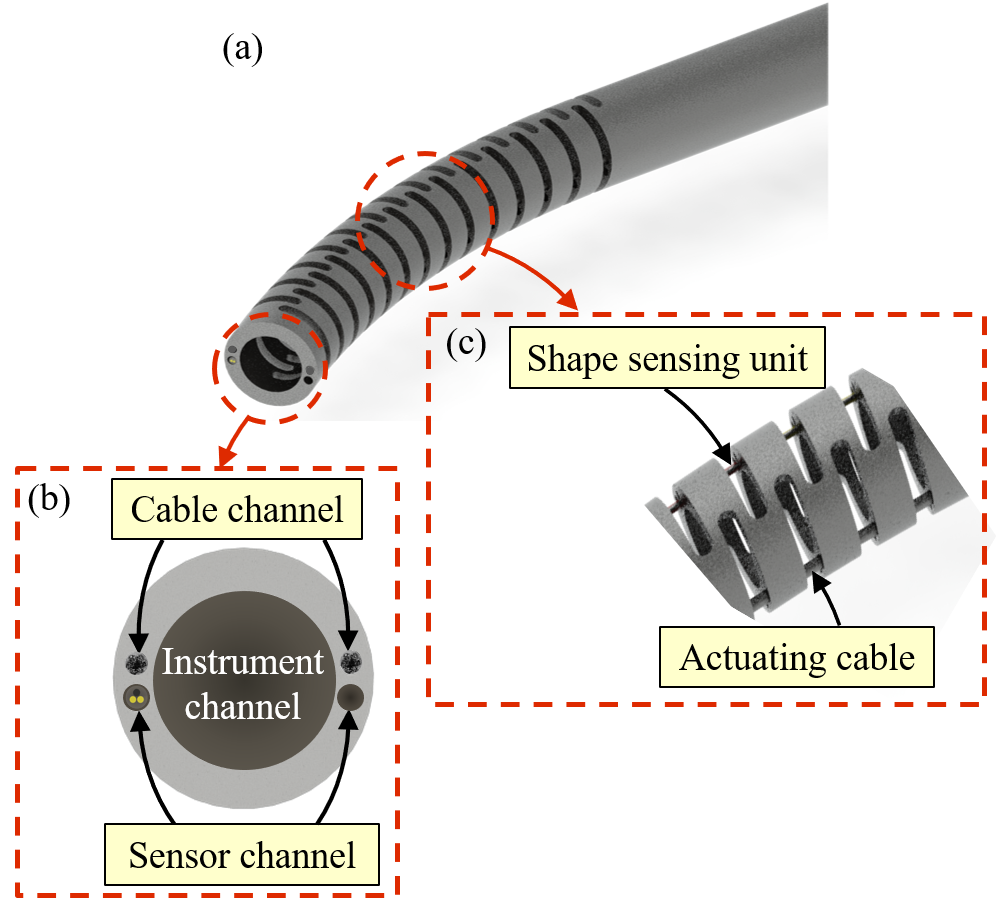}
\caption{{\color{subsectioncolor}{(a)}} Schematics of the CDM while bending. {\color{subsectioncolor}{(b)}} Cross-section of the CDM tip, including sensor, cable, and instrument lengthwise channels. {\color{subsectioncolor}{(b)}} Detailed view of the CDM flexible segment with the shape sensing unit and the actuating cable.}
\label{CDM_Sensor_Cable}
\end{figure}

As shown in {\color{subsectioncolor}{Fig.\ref{CDM_Sensor_Cable}a}}, the cable-driven CDM employed in the orthopaedic procedures has been constructed from a Nitinol (NiTi) rod with an outside diameter (OD) of 6 mm and several notches along its 35 mm flexible length to reach compliance in the bending direction while achieving high stiffness in the perpendicular direction to the bending plane \cite{murphy2014design}. The CDM also consists of a 4 mm diameter open lumen as an instrument channel for passing flexible debriding tools\cite{alambeigi2017curved} ({\color{subsectioncolor}{Fig.\ref{CDM_Sensor_Cable}b}}). The overall length of the CDM is 70 mm, and the CDM wall contains four lengthwise channels such that each pair of channels is along the two opposite sides of the wall ({\color{subsectioncolor}{Fig.\ref{CDM_Sensor_Cable}b}}). As shown in {\color{subsectioncolor}{Fig.\ref{CDM_Sensor_Cable}c}}, the actuating cable is passed through its channel with a 0.5 mm diameter to create the bidirectional planar bending; the sensor channel with a diameter of 0.6 mm is for embedding the FBG-based shape sensing unit into the CDM and tracking the 2D-shape of the CDM in real-time\cite{sefati2020surgical}.

The design requirements of the sensor assembly are derived from the CDM specifications and its applications. For orthopaedic applications such as core decompression of the hip \cite{alambeigi2017curved} and the treatment of osteolysis behind the acetabular implant \cite{sefati2020surgical}, the outer diameter of the CDM must be less than 6 mm. The CDM has, therefore, a thin wall thickness (1 mm), with sensor channels of less than 0.5 mm \cite{sefati2021dexterous}. Considering the major challenges of the previous work as well as the CDM design constraints, the development of an FBG-based shape sensing unit needs to meet some crucial requirements: (1) easy fabrication and integration into the CDM; (2) inexpensive compared to multi-core optical fibers; (3) high wavelength shift to curvature ratio (sensitivity) for the small local deformations, (4) measure high deformations without surpassing the FBG strain limit, (5) less than 0.6 mm diameter to ensure the easy sliding of the shape sensing unit through the sensor channel, and (6) an enclosed structure to protect optical fibers and prevent adhesive wear due to the sensing unit slide inside the sensor channel.

%The design requirement of the sensor assembly is derived from the CDM specifications. The dimensional constraints of the CDM, such as the screw hole diameter of the acetabular implant and the existence of the instrument channel, lead to a thin CDM wall thickness (1 \textit{mm}), and thus limitation in the diameter of the sensor channel\cite{sefati2020surgical}. 

\subsection{FBG Sensor Configuration}
\begin{figure}[!t]
\centering
\includegraphics[width=\columnwidth,height=2.2in]{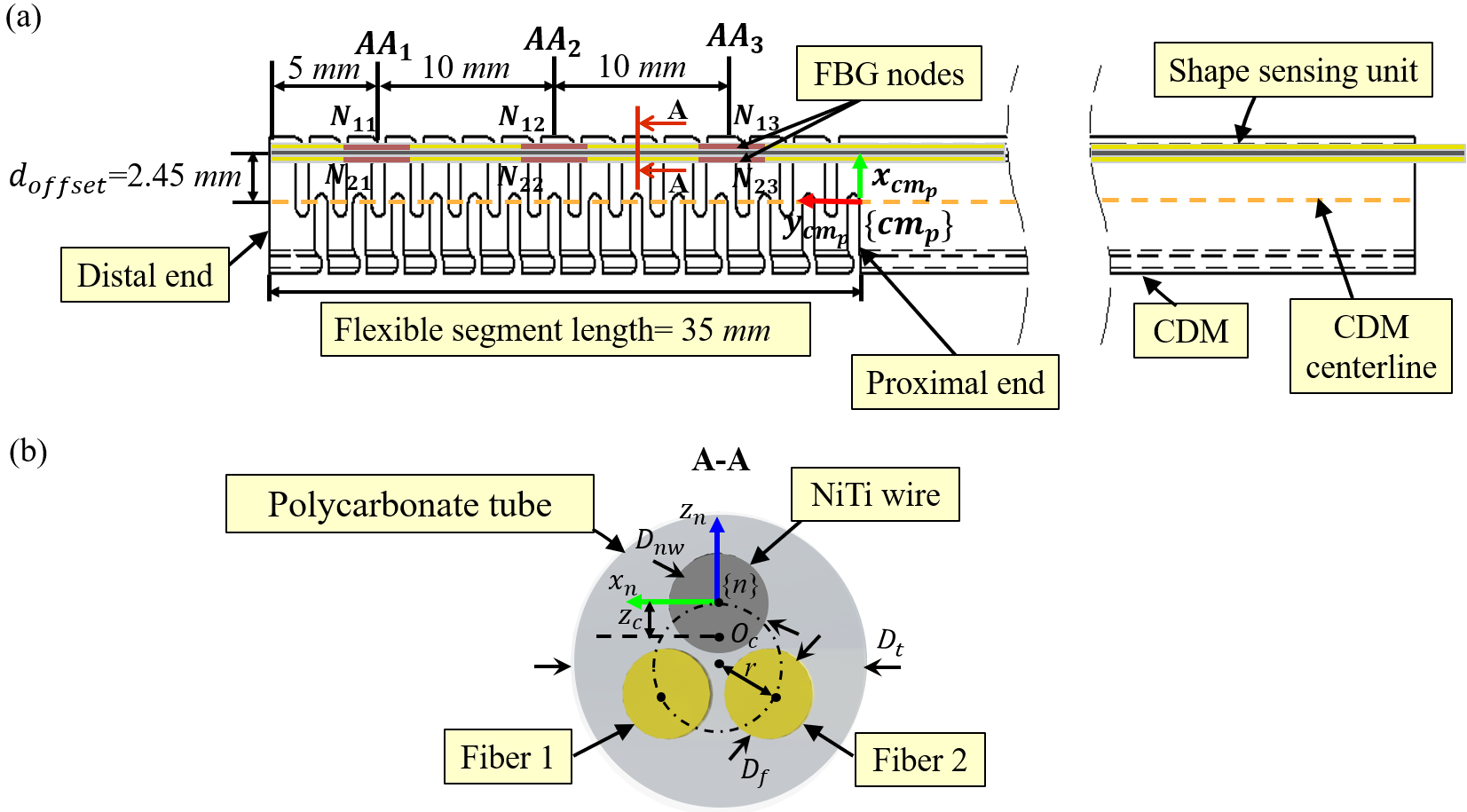}
\caption{{\color{subsectioncolor}{(a)}} Arrangement of FBG nodes, including two fibers and one NiTi rod, inside the CDM. Three sets of active areas along the length of the sensing unit are denoted by: ${AA}_1$, ${AA}_2$, ${AA}_3$. {\color{subsectioncolor}{(b)}} Shape sensing unit in the cross-section view. The two cores of fibers are labeled by Fiber1 and Fiber2.}
\label{FBG_Arrangement_CDM}
\end{figure}

To address the aforementioned requirements, a polycarbonate tube with an outer diameter of 500\textpm15 \textit{\si{\micro\meter}} and three symmetric 150\textpm5 \textit{\si{\micro\meter}} lumens was developed (Paradigm Optics, Inc.)\cite{sefati2016fbg}. Three lumens are 120$^{\circ}$ radially apart on a circle that has a 100 \textit{\si{\micro\meter}} diameter. The three-lumen polycarbonate tube as a substrate meets requirements 1 and 5 to ensure the simple sensor assembly, consistently repeatable fabrication, and unchallenging placement of the sensing unit into the CDM. The flexible enclosed structure of the polycarbonate tube fully covers optical fibers, which fulfills requirement 6. The strain of FBG fibers should not exceed their allowable range. The sensing unit can achieve high sensitivity as well as large deformations if the orthogonal distance between the FBG sensor and the neutral axis of the sensor assembly, called sensor orthogonal distance, is neither too large nor too small\cite{jang2019ultra,ge2016bidirectional} (requirements 3 and 4). Hence, our sensing unit is designed with two optical fibers and one NiTi rod with a 125 \textit{\si{\micro\meter}} diameter ({\color{subsectioncolor}{Fig.\ref{FBG_Arrangement_CDM}b}}). NiTi rod is chosen due to its super elastic property, and the capability of undergoing large deflections. Also, the NiTi rod can increase the rigidity of the sensor assembly and, hence, its sensitivity.
An optical fiber with an 80 \textit{\si{\micro\meter}} cladding diameter and a 120 \textit{\si{\micro\meter}} coating diameter (Technica Optical Components, LLC., Atlanta, GA) offers a small size for insertion into the three-lumen polycarbonate tube. The strain range of the optical fiber is up to 1.5\%. {\color{subsectioncolor}{Fig.\ref{FBG_Arrangement_CDM}a}} shows the arrangement of FBG nodes inside the CDM. The sensing unit is along the wall of the sensor channel which is $d_{offset}=$ 2.45 mm away from the CDM centerline. Since each fiber consists of three FBG nodes with a 5 mm length, the sensing unit includes three active areas (${AA}_1-{AA}_3$) in total. ${N}_{kj}$ is the $j^{th}$ FBG node on the fiber $k\in\{1,2\}$ and ${AA}_j$ is the $j^{th}$ active area. $j\in\{1,2,3\}$ is the subscript that indicates the number associated with the active area as well as the corresponding FBG nodes. The center of the active areas are 10 mm apart, and the distance between the center of the active area ${AA}_1$ and the distal end of the CDM is 5 mm. This arrangement of active areas provides adequate coverage for reconstructing the CDM centerline.
%$z_{c}\in{R_{>0}}$ is the distance from the center of the NiTi wire to the geometric centroid of the sensing unit; $r\in{R_{>0}}$ is the radius of the circle on which the centers of two fibers and the NiTi wire are located.
%Specifically, in the case of increase in the distance between the FBG arrays and the sensor assembly’s neutral plane, the sensing unit sensitivity and applied strain increase which leads to reaching the FBG strain limit at a small local deformation\cite{ge2016bidirectional}.
\subsection{Neutral Plane of the Sensing Unit}
The neutral axis is normally positioned at the location of the geometric centroid. We assume that the sensing unit is a composite beam consisting of different materials at each cross-section. As shown in {\color{subsectioncolor}{Fig.\ref{FBG_Arrangement_CDM}b}}, the sensing unit cross-section is symmetric about the $z_{n}$-axis of the local frame \{n\}, which is located at the center of the NiTi rod. Since the neutral plane always passes through the geometric centroid of the sensor assembly, $O_{c}$, the \textit{y}-coordinate of the geometric centroid is zero, $[y_{c}]_{\{n\}}=0$, at each cross section. The \textit{z}-coordinate of the geometric centroid, $z_{c}\in R^{>0}$, can be derived from the equilibrium equation of forces in the local frame \{n\}:
\begin{equation}
\label{neutralaxis}
\sum_{i=1}^{4} \int_{A_i} \sigma_i dA_i=0,\: \sigma=\frac{E_i(z-z_{c})}{\rho}.
\end{equation}

where $\rho$ corresponds to the radius of the curvature. The \textit{i} subscript indicates the components of the sensing unit including the polycarbonate tube, NiTi rod, FBG fibers, and lumens of the polycarbonate tube. Also, $\sigma_i$, $E_i$, and $A_i$ are the stress, Young’s modulus, and the cross-section area of the $i^{th}$ component of the sensor assembly, respectively. Using \eqref{neutralaxis}, the $z_{c}$ in the local frame \{n\} can be obtained by:
\begin{equation}
\label{z-coordinate-of-neutralaxis}
[z_{c}]_{\{n\}}=\frac{(3E_f D_f^2+E_t (D_t^2-3D_l^2 ))r}{E_{nw} D_{nw}^2+2E_f D_f^2+E_t (D_t^2-3D_l^2 ) },
\end{equation}

where components of the sensor assembly namely polycarbonate tube, NiTi rod, FBG fibers, and lumens of the polycarbonate tube are represented as subscripts \textit{t}, \textit{nw}, \textit{f}, and \textit{l}, respectively. \textit{D} denotes the diameter of the sensor assembly component and $r\in R^{>0}$ denotes the radius of the circle in which the center of three lumens of the polycarbonate tube is located. The Young’s modulus of the polycarbonate tube, NiTi rod and FBG fiber are 2.6 \textit{GPa}, 75 \textit{GPa}, and 70 \textit{GPa}, respectively. By substituting the properties and dimensions of the sensor assembly components into \eqref{z-coordinate-of-neutralaxis}, the value of the $[z_{c}]_{\{n\}}$ becomes 0.095 mm.

%Changing the bending direction of the sensing unit from 0$^{\circ}$ to 180$^{\circ}$ causes a decreasing trend in the sensor orthogonal distance and the sensitivity of the adjacent fiber to the neutral axis, and an increasing one in the distant fiber to the neutral axis. 
Changing the bending direction of the sensing unit affects the sensor's orthogonal distance and hence the sensor sensitivity. To satisfy requirements 3 and 4 for both optical fibers, the sensing unit's neutral plane should be perpendicular to the CDM bending plane, making the sensor's orthogonal distance to the two fibers equal.

\subsection{FBG Sensor Assembly}
\begin{figure}[!t]
\centering
\includegraphics[width=\columnwidth]{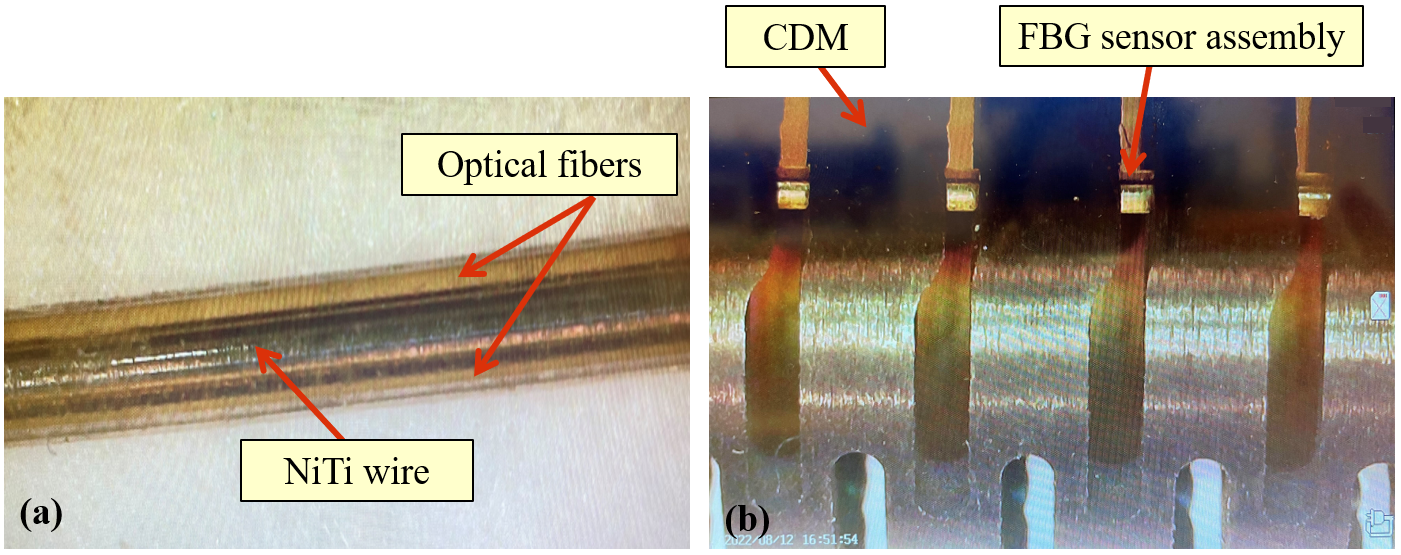}
\caption{{\color{subsectioncolor}{(a)}} Section of the FBG sensor assembly. {\color{subsectioncolor}{(b)}} FBG sensor assembly embedded in the CDM.}
\label{FBG_under_Microscope}
\end{figure}

Since the core of the fiber alone is not affected by the bending strain, using the polycarbonate tube as a substrate can provide an offset between the center of the fiber and the neutral axis. Also, the fixed geometry of the three-lumen polycarbonate tube can help with the precise placement of sensor components. Two fibers were first passed through the two lumens while the tube was kept straight. After embedding and aligning both fibers in the longitudinal direction of the tube, the UV glue, which has a low viscosity, was passed through the lumens by submerging one end in the UV glue and applying suction to the other end.
%was injected into the lumens by a suction machine. 
The glue injection was continued until all trapped air bubbles which are a potential source of the sensor error were removed. The NiTi rod was then passed through the third lumen, and finally, the UV glue inside the lumens was cured using a UV spot gun. {\color{subsectioncolor}{Fig.\ref{FBG_under_Microscope}a}} illustrates the segment of the shape sensing unit after assembly. The new approach makes the sensor fabrication process far easier, less time-consuming ( 4 hours of assembly), and low-cost (requirement 2) compared to previously reported designs \cite{liu2015large,sefati2017highly,khan2019multi}. 

After assembly, as described in section II-C, the sensing unit is passed through the CDM sensor channel such that its neutral plane is kept perpendicular to the CDM’s bending plane ({\color{subsectioncolor}{Fig.\ref{FBG_under_Microscope}b}}). The sensor tip is then glued at the distal end of the CDM. The sensor, hence, can move freely as the CDM bends or straightens.

\section{methods}
\subsection{The Principal of FBG-based Shape Sensing}
The FBG wavelength shift is linearly dependent on the mechanical strain, $\varepsilon$, and the temperature,  T, which is given by\cite{abayazid20133d,roesthuis2013three}:
\begin{equation}
\label{FBG-principal}
\frac{\Delta\lambda_B}{\lambda_{B_0 }} =K_{\varepsilon} (\varepsilon-\varepsilon_{0})+K_T (T-T_0),
\end{equation}

where $\Delta\lambda_B$ is the FBG wavelength shift and $\lambda_{B_0 }$ is the reference FBG wavelength at the reference strain, $\varepsilon_{0}$, and the reference temperature, $T_0$. $K_{\varepsilon}$ and $K_T$ refer to the constant coefficients associated with the strain and the temperature, respectively. The strain coefficient is directly related to the photo-elastic coefficient by $K_{\varepsilon}=1-p_{\varepsilon}$, where the photo-elastic coefficient is set to 0.22 \cite{li2018sensitivity}. In this study, the reference wavelength is collected when the fiber is straight, the reference strain, thus, can be assumed to be zero. Also, due to the proximity of fibers, the temperature variation of FBG nodes at each active area is the same \cite{xiong2019investigation}. In the case of the constant temperature, by measuring the FBG wavelength, strain can be calculated using \eqref{FBG-principal}.

\subsection{Curvature Calculation of the Shape Sensor}
\begin{figure}[!t]
\centering
\includegraphics[width=\columnwidth]{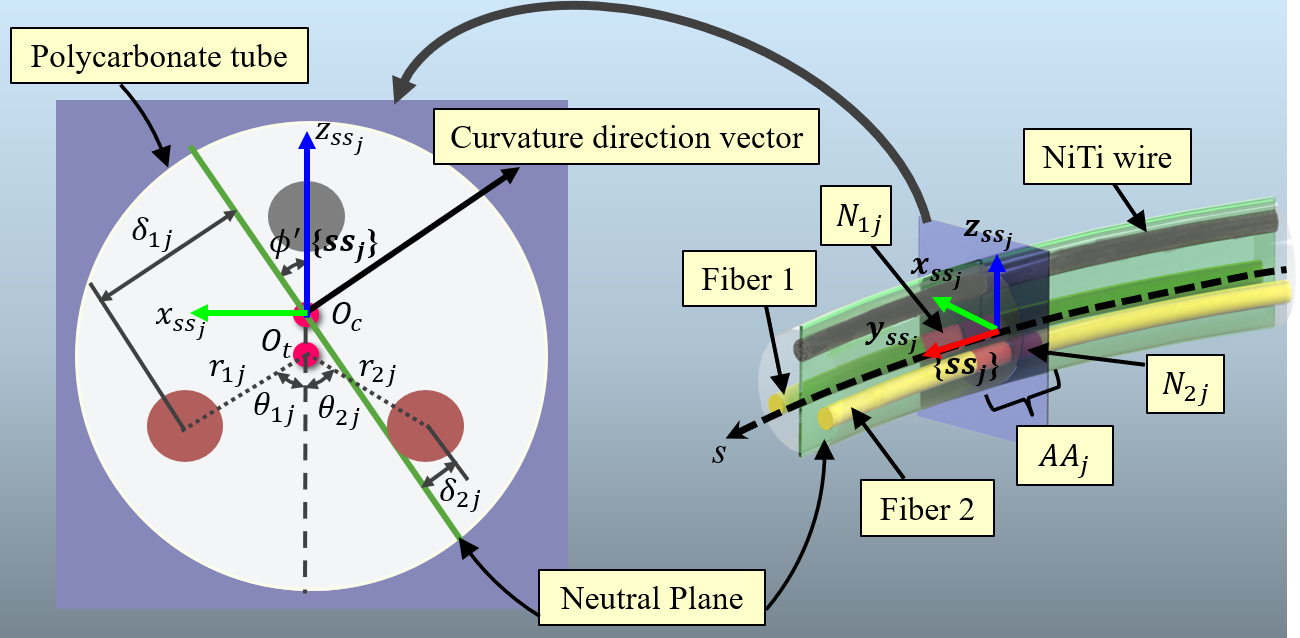}
\caption{Section of the shape sensing unit at ${AA}_j$ with the cross-section view at FBG nodes labeled by $N_{1j}$ and $N_{2j}$.}
\label{Section_of_SS}
\end{figure}

The curvature and the bending direction of the shape sensor can be obtained by strain calculation at each active area, referred to ${AA}_j$. {\color{subsectioncolor}{Fig.\ref{Section_of_SS}}} illustrates the cross-section of the shape sensing unit at ${AA}_j$. \{$ss_{j}$\} is the frame of the shape sensor centerline at the $j^{th}$ active area. The location of each active area and its associated strain, curvature, and bending direction are parameterized by the arc length, $s$, which passes through the geometric centroid of the sensing unit. The arc length parameter is for reconstructing the shape sensor centerline. Axis $y_{ss_J}$ is tangential to the shape sensor centerline, and the direction of the axis $x_{ss_J}$ is from the origin of the frame \{$ss_{j}$\} to the fiber 1 at the $j^{th}$ active area. Considering the pure bending for the sensing unit, equations which relate the axial strain, $\varepsilon_{kj}$, of the $j^{th}$ FBG node on the fiber $k\in \{1,2\}$ to the corresponding curvature and bending direction are as follows\cite{roesthuis2013using}:
\begin{equation}
\label{1-FBG-strain-tempreture}
    \begin{aligned}
    \varepsilon_{1j} (s) &=-\kappa^{'}(s) \delta_{1j} \\
      &=-\kappa^{'}(s)[z_{c} \text{sin}(\phi^{'}(s))-r_{1j} \text{sin}(\theta_{1j} (s)-\phi^{'}(s))],
    \end{aligned}
\end{equation}
\begin{equation}
\label{2-FBG-strain-tempreture}
    \begin{aligned}
    \varepsilon_{2j} (s) &=-\kappa^{'}(s) \delta_{2j} \\
      &=-\kappa^{'}(s)[z_{c} \text{sin}(\phi^{'}(s))+r_{2j} \text{sin}(\theta_{2j} (s)+\phi^{'}(s))].
    \end{aligned}
\end{equation}
 
 where $\phi^{'}$ indicates the bending direction which is the angle between the axis $z_{ss_{j}}$ of the local frame \{$ss_{j}$\} and the neutral axis. $\kappa^{'}$ is the curvature, $\delta_{kj}$ denotes the sensor orthogonal distance of the $j^{th}$ FBG node on the $k^{th}$ fiber, $r_{kj}$ denotes the radial distance from the $j^{th}$ FBG node on the $k^{th}$ fiber to the center of the polycarbonate tube, $\theta_{kj}$ is the angular offset from the negative axis $z_{ss_{j}}$ to the core of $j^{th}$ FBG node on the $k^{th}$ fiber.
 
By assuming the reference strain to be zero, the following sets of equations can be derived from substituting \eqref{1-FBG-strain-tempreture} and \eqref{2-FBG-strain-tempreture} into \eqref{FBG-principal}:
%and the temperature variation of FBG nodes at each active area is constant%
\begin{equation}
\label{1-Simplified-FBG-strain-tempreture}
    \begin{aligned}
    \frac{\Delta\lambda_{1j}}{\lambda_{0_{1j}}} =& -K_{\varepsilon} \kappa^{'}(s)[z_{c} \text{sin}(\phi^{'}(s)) \\
      & -r_{1j}  \text{sin}(\theta_{1j} (s)-\phi^{'}(s))]+K_T \Delta T_j,
    \end{aligned}
\end{equation}
\begin{equation}
\label{2-Simplified-FBG-strain-tempreture}
    \begin{aligned}
    \frac{\Delta\lambda_{2j}}{\lambda_{0_{2j}}} =& -K_{\varepsilon} \kappa^{'}(s)[z_{c} \text{sin}(\phi^{'}(s)) \\
      & +r_{2j}  \text{sin}(\theta_{2j} (s)+\phi^{'}(s))]+K_T \Delta T_j.
    \end{aligned}
\end{equation}

where the term $\Delta T_j$ is the temperature variation at the $j^{th}$ active area. It is assumed that the term $K_T \Delta T_j$ is the same for both fibers due to the proximity of the FBG nodes at each active area. $\Delta\lambda_{kj}$ and $\lambda_{0_{kj}}$ are the wavelength shift and the reference wavelength associated to the $j^{th}$ FBG node on the $k^{th}$ fiber, respectively. Equations \eqref{1-Simplified-FBG-strain-tempreture} and \eqref{2-Simplified-FBG-strain-tempreture} can be simplified and then written in a matrix form:
\begin{equation}
\label{Matrix-FBG-strain-tempreture}
    \begin{aligned}
    \Lambda_{j}=A_j B_j,
    \end{aligned}
\end{equation}

where:
\begin{gather*}
\Lambda_{j}= 
\begin{bmatrix}
\frac{\Delta\lambda_{1j}}{\lambda_{0_{1j}}} \\
\frac{\Delta\lambda_{2j}}{\lambda_{0_{2j}}}
\end{bmatrix}
, B_{j}=
\begin{bmatrix}
B_{j_1} \\
B_{j_2} \\
B_{j_3}
\end{bmatrix}
=
\begin{bmatrix}
\kappa^{'}(s)\text{sin}(\phi^{'}(s) \\
\kappa^{'}(s)\text{cos}(\phi^{'}(s) \\
K_T \Delta T_j
\end{bmatrix}\\
A_j=
\begin{bmatrix}
-K_{\varepsilon} (z_{c}+r_{1j}  \text{cos}(\theta_{1j} (s)) ) & K_{\varepsilon} r_{1j}  \text{sin}(\theta_{1j} (s)) & 1\\
-K_{\varepsilon} (z_{c}+r_{2j}  \text{cos}(\theta_{2j} (s)) ) & -K_{\varepsilon} r_{2j}  \text{sin}(\theta_{2j} (s)) & 1
\end{bmatrix}
.
\end{gather*}

$B_j$ from \eqref{Matrix-FBG-strain-tempreture} can be obtained as:
\begin{equation}
\label{B-Matrix-FBG-strain-tempreture}
    \begin{aligned}
    B_j=A_j^{\dag} \Lambda_j,
    \end{aligned}
\end{equation}

where $\dag$ is the Moore-Penrose pseudo-inverse symbol. By solving \eqref{B-Matrix-FBG-strain-tempreture}, the curvature and the bending direction at the $j^{th}$ active area can be determined as:
\begin{equation}
\label{curvature}
    \begin{aligned}
    \kappa^{'}(s)=\sqrt{(B_{j_{1}}^2+B_{j_{2}}^2 )},
    \end{aligned}
\end{equation}
\begin{equation}
\label{curvature direction}
    \begin{aligned}
    \phi^{'}(s)=\text{atan2}(B_{j_1 },B_{j_2 }).
    \end{aligned}
\end{equation}

\begin{figure}[!t]
\centering
\includegraphics[width=\columnwidth]{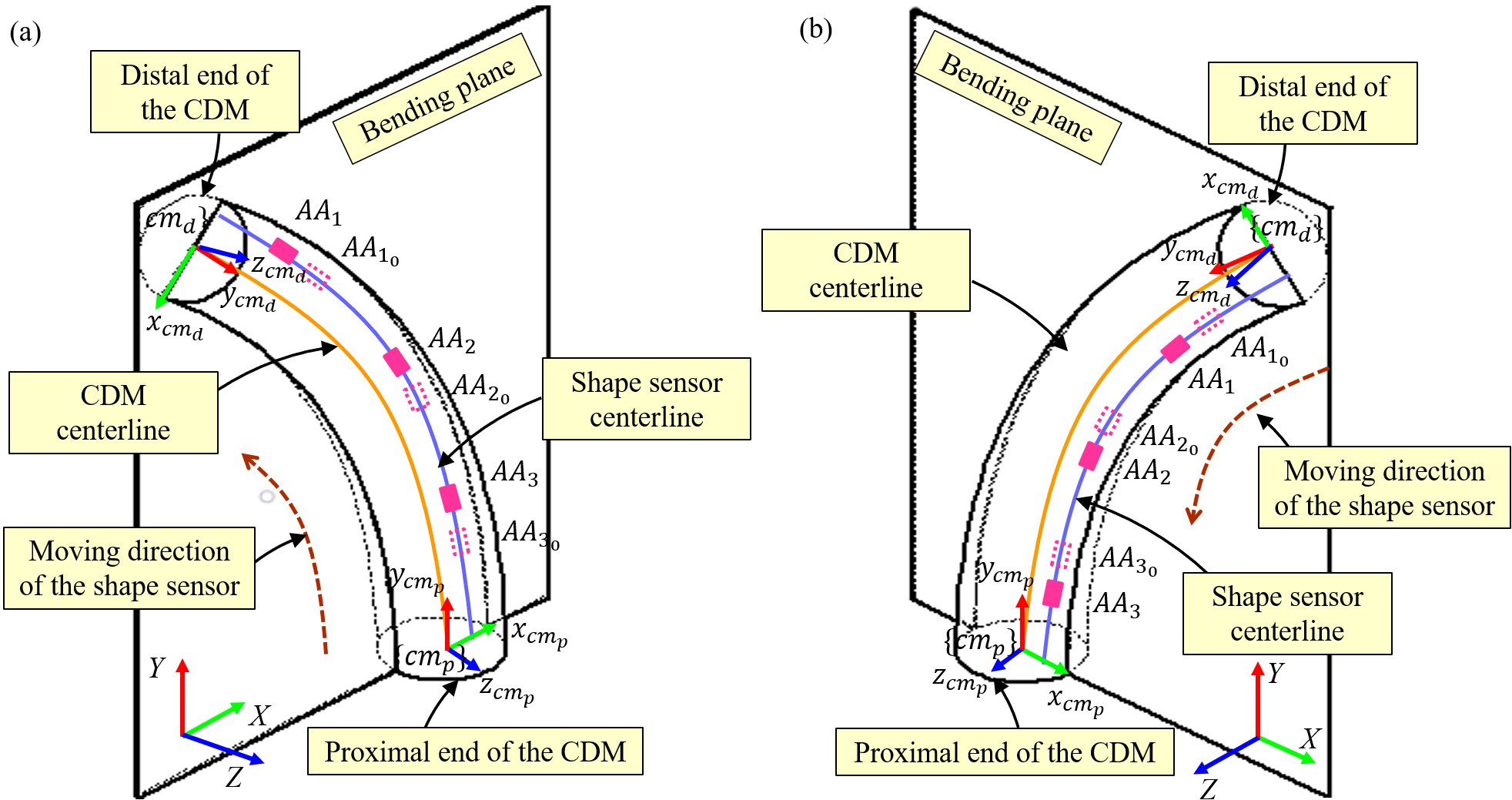}
\caption{Moving direction of the shape sensing unit at {\color{subsectioncolor}{(a)}} CDM negative deflection and {\color{subsectioncolor}{(b)}} CDM positive deflection. ${AA}_{1_{0}}$, ${AA}_{2_{0}}$, and ${AA}_{3_{0}}$ are the initial locations of the active areas when the CDM is straight and ${AA}_1$, ${AA}_2$, and ${AA}_3$ are the active areas locations after the CDM bending.}
\label{FBG_Sliding}
\end{figure}

 As shown in {\color{subsectioncolor}{Fig.\ref{FBG_Sliding}}}, the CDM can only bend in the \textit{XY} plane of the planar bending which is aligned with the frame \{$cm_{p}$\}. \{$cm_{p}$\} is the frame of the CDM centerline at the proximal end of the CDM. Axis $y_{cm_p}$ is tangential to the CDM centerline, and the direction of the axis $x_{cm_p}$ is from the origin of the frame \{$cm_{p}$\} to the sensing unit. The wavelength shift sign of the sensing unit can simply specify the moving direction of the sensing unit as well as the CDM deflection direction; if the wavelength shifts of the FBG nodes on fiber 1 and fiber 2 are positive and negative, respectively, the CDM is undergoing positive deflection and the sensing unit is moving forward to the distal end of the CDM, if the situation is opposite, the CDM is undergoing negative deflection and the sensing unit is moving back to the proximal end of the CDM. Of note, the friction between the shape sensor and the wall of the CDM sensor channel affects the FBG wavelength measurements. To compensate for the influence of the friction, we consider a calibration coefficient at the $j^{th}$ active area, represented by $C_j$:
\begin{equation}
\label{curvature correction}
    \begin{aligned}
    \kappa[j]=C_j \kappa^{'}[j],\quad j=1, 2, 3
    \end{aligned}
\end{equation}

where $\kappa[j]$ is the compensated curvature at the $j^{th}$ active area after applying the calibration coefficient, and $\kappa^{'}[j]$ is the curvature at the $j^{th}$ active area before applying the calibration coefficient. Furthermore, the sensing unit may exhibit internal twists which result in the out-of-plane CDM centerline reconstruction. The effect of the twist on the bending direction is compensated at each active area by:
\begin{equation}
\label{twist compensation}
    \begin{aligned}
    \phi[j]=\phi^{'}[j]-\phi_{t}[j], \quad j=1, 2, 3
    \end{aligned}
\end{equation}

where $\phi[j]$ is the compensated bending direction at the $j^{th}$ active area, and $\phi_{t}[j]$ is the twisted bending direction at the $j^{th}$ active area. Two steps are required to find $\phi_{t}[j]$: 1) the shape sensing unit is lying straight in the \textit{XY} plane of the bending plane such that the NiTi rod is facing the positive \textit{Z} direction. The wavelengths at each active area are recorded as the reference wavelength, $\lambda_{0_{kj}}$; 2) the shape sensing unit is inserted in a constant curvature groove which implies bending in the \textit{XY} plane of the bending plane. Then, the twisted bending direction at each active area due to the internal twist can be computed using \eqref{B-Matrix-FBG-strain-tempreture} and \eqref{curvature direction}. These values are constant and do not change during the bending of the CDM, embedded with the sensing unit. 
%Since the CDM can only perform the bidirectional planar bending, the values of the compensated curvature and the bending direction are sufficient to reconstruct the 2D-shape of the sensing unit.%
\subsection{Sensor Assembly Shape Reconstruction}
\begin{figure}[!t]
\centering
\includegraphics[width=2.4in]{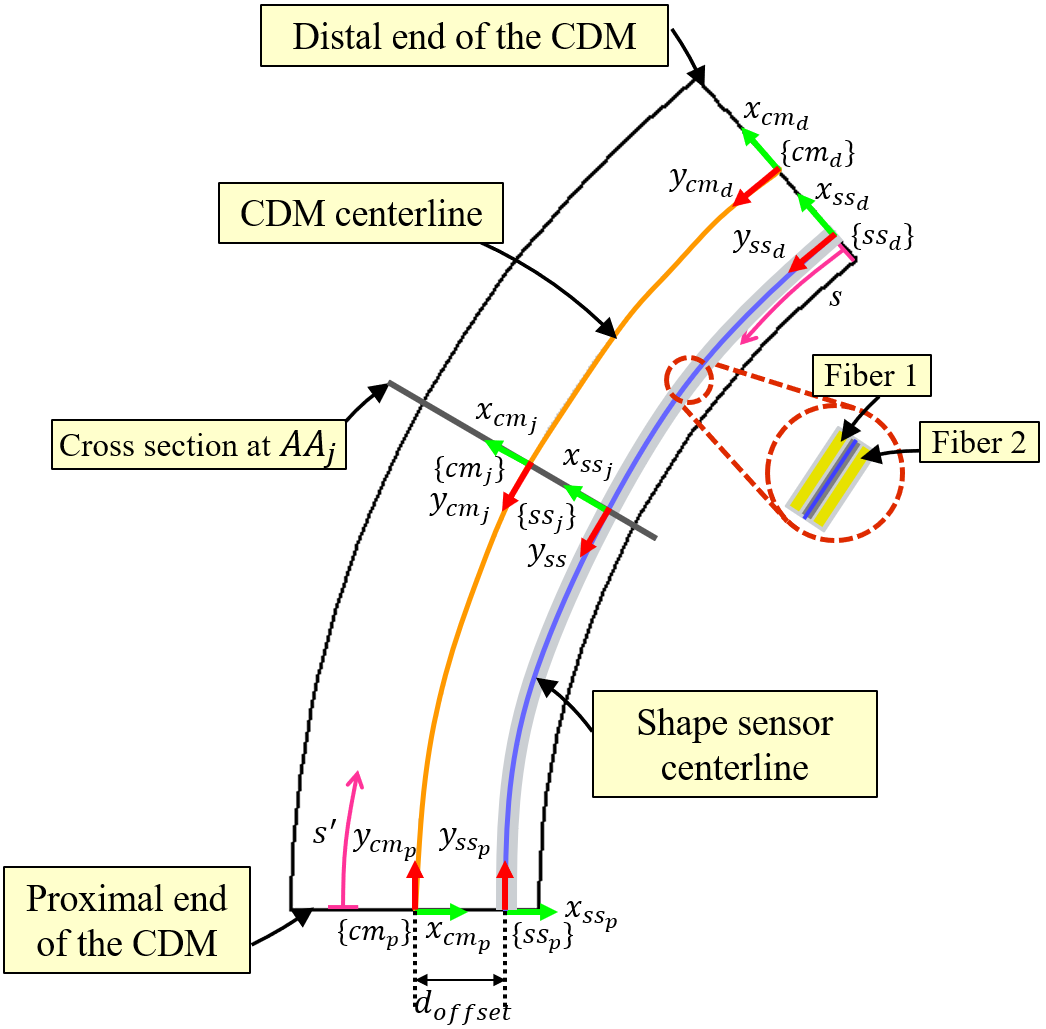}
\caption{The side cross-section view of the CDM for shape reconstruction of the sensor assembly and the CDM. $\{{ss}_p\}$ and $\{{ss}_d\}$ are the frames of the shape sensor centerline at proximal and distal ends, respectively. Axes $\{y_{{ss}_p}\}$ and $\{y_{{ss}_d}\}$ are tangential to the curve of the shape sensor centerline. The direction of axes $\{x_{{ss}_p}\}$ and $\{x_{{ss}_d}\}$ are from the origin of their associated frames to the fiber 2 and fiber 1, respectively.}
\label{CDM-centerline-reconstruction}
\end{figure}
%it is impossible to find the arc length of each active area from the proximal end. Hence,%
As described in section III-B, the sensor assembly moves freely, while CDM bends ({\color{subsectioncolor}{Fig.\ref{FBG_Sliding}}}). Since the sliding amount of the sensing unit is unknown, the shape of the sensing unit is reconstructed from the distal end rather than the proximal end, where the arc lengths of all active areas remain constant. As shown in {\color{subsectioncolor}{Fig.\ref{CDM-centerline-reconstruction}}}, some coordinate systems are defined for reconstructing the shape sensor centerline.  The measured FBG wavelengths are utilized to determine the curvature and the bending direction at the location of each active area, using \eqref{curvature}-\eqref{twist compensation}. By knowing the arc length of each active area in frame $\{{ss}_d\}$ as well as the corresponding curvature and bending direction, spline interpolation can be performed to compute the continuous curvature, $\kappa(s)$, and the bending direction, $\phi(s)$. The interval of the arc length of the shape sensor centerline which is denoted by $s$, is from the origin of the frame $\{{ss}_d\}$ to the last active area, $AA_3$.

\begin{figure}[!t]
\centering
\includegraphics[width=2.7in]{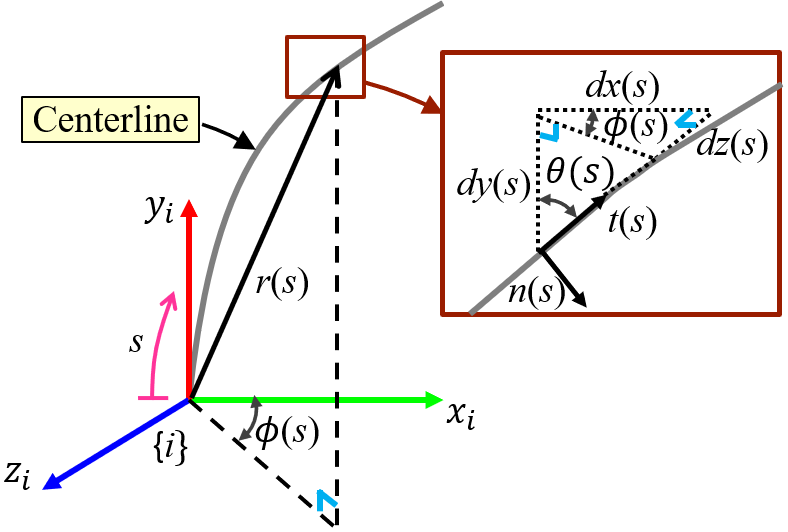}
\caption{Centerline representation which is parameterized by a planar curve, $r(s)$, with Frenet-Serret vectors including the tangent vector, $t(s)$, and the normal vector, $n(s)$.}
\label{Frenet-eq}
\end{figure}

As shown in {\color{subsectioncolor}{Fig.\ref{Frenet-eq}}}, the centerline can be parametrized by the arc length as $r(s)$ in the frame $\{i\}$. Using the Frenet-Serret apparatus, the tangent of the curve, $t(s)$, is given by:
\begin{equation}
\label{tengent eq}
    \begin{aligned}
    [t(s)]_{\{i\}}=\frac{dr(s)}{ds}=
    \begin{bmatrix}
    \frac{dx(s)}{ds} & \frac{dy(s)}{ds} & 0 
    \end{bmatrix}
    ^{T},
    \end{aligned}
\end{equation}

where \textit{x(s)} and \textit{y(s)} represent the 2D position of a point along the centerline in the frame $\{i\}$. For the sensor assembly shape reconstruction, frame $\{{ss}_d\}$ is considered as the frame $\{i\}$. The slope of the centerline is related to the curvature by:
\begin{equation}
\label{theta-shape-sensor}
    \begin{aligned}
    \theta(s)=\int_0^s \kappa(s) ds+\theta_0 ,
    \end{aligned}
\end{equation}

where $\theta(s)$ is the slope of the centerline which is the angle between the tangent vector at a point along the centerline and the positive direction of the $y_{\{i\}}$-axis in the frame $\{i\}$. $\theta_0$ is the initial value of the slope of the centerline at \textit{s=0}. By taking the integral of $t(s)$ in \eqref{tengent eq}, \textit{x} and \textit{y} positions of the centerline in term of the arc length can be derived as:
\begin{equation}
\label{curve of the shape sensor}
    \begin{aligned}
    [r(s)]_{\{i\}}=
    \begin{bmatrix}
    (\int_0^s \text{sin}(\theta(s)) ds+x_0)\text{cos}(\phi(s)) \\
    \int_0^s \text{cos}(\theta(s)) ds+y_0 \\
    0
    \end{bmatrix}
    ,
    \end{aligned}
\end{equation}

where $x_0$ and $y_0$ are the initial deflections in the \textit{x} and \textit{y} directions, respectively. Finally, the coordinates of the shape sensor centerline in the frame $\{{ss}_d\}$ can be obtained using \eqref{theta-shape-sensor} and \eqref{curve of the shape sensor}.

\subsection{CDM Shape Reconstruction}
The shape sensor centerline obtained in section III-C should be translated to the CDM centerline, which has a constant offset, $d_{offset}$, from the shape sensor centerline ({\color{subsectioncolor}{Fig.\ref{CDM-centerline-reconstruction}}}). $\{{cm}_d\}$ and $\{{cm}_j\}$ are the frames of the CDM centerline at the distal end and the corresponding cross-section of the $j^{th}$ active area, respectively. Axes $y_{{cm}_j}$ and $y_{{cm}_d}$ are tangential to the curve of the CDM centerline. The direction of axes $x_{{cm}_j}$ and $x_{{cm}_d}$ are the same as axes $x_{{ss}_j}$ and $x_{{ss}_d}$, respectively. The coordinate of the $j^{th}$ active area on the CDM centerline in the frame $\{{cm}_d\}$ is determined by:
\begin{equation}
\label{CDM-coordinate-in-D}
    \begin{aligned}
    \begin{bmatrix}
    x_{O_{cm_{j}}}\\
    y_{O_{cm_{j}}}\\
    0\\
    1
    \end{bmatrix}_{\{{cm}_d\}}
    =
    \begin{bmatrix}
    R^{{ss}_d}_{{ss}_j}(\theta[j]) & %P^{{ss}_d}_{{ss}_j}+P^{{cm}_d}_{{ss}_d}\\
    V^{{cm}_d}_{{ss}_j}\\
    [0 \quad 0 \quad 0] & 1
    \end{bmatrix}
    \begin{bmatrix}
    x_{O_{cm_{j}}}\\
    y_{O_{cm_{j}}}\\
    0\\
    1
    \end{bmatrix}_{\{{ss}_j\}}
    ,
    \end{aligned}
\end{equation}

where $x_{O_{cm_{j}}}$ and $y_{O_{cm_{j}}}$ denote the coordinates of the origin of the frame $\{cm_{j}\}$, and $R^{{ss}_d}_{{ss}_j}$ is a 3$\times$3 rotation matrix about the \textit{z}-axis from the frame $\{{ss}_d\}$ to the frame $\{{ss}_j\}$ by an angle $\theta[j]$. $V^{{cm}_d}_{{ss}_j}$ which is a 3$\times$1 translation vector from the frame $\{{cm}_d\}$ to the frame $\{{ss}_j\}$ is found using \eqref{Translation-vector-cmd}:
\begin{equation}
\label{Translation-vector-cmd}
    \begin{aligned}
    V^{{cm}_d}_{{ss}_j}
    =
    [x[j]-d_{offset} \quad y[j] \quad 0]^T_{\{cm_d\}},
    \end{aligned}
\end{equation}
%\begin{equation}
%\label{Translation-vector-cmd}
 %   \begin{aligned}
 %   P^{{cm}_d}_{{ss}_d}
 %   =
 %   [-d_{offset} \quad 0 \quad 0]^T_{\{cm_d\}}
%    \end{aligned}
%\end{equation}

where $x[j]$ and $y[j]$ are the coordinates of the $j^{th}$ active area on the shape sensor centerline. After finding the coordiantes of active areas on the CDM centerline in the frame $\{{cm}_d\}$, it is possible to perform spline interpolation to find the arc length of each active area in the frame $\{{cm}_p\}$:
\begin{equation}
\label{arc-length-in-P}
    \begin{aligned}
    s^{'}[j]=L-\int_0^{x_{O_{cm_{j}}}} \sqrt{1+(\frac{df(x_{cm})}{dx_{cm}})^2} \, dx_{cm},
    \end{aligned}
\end{equation}

where $f(x_{cm})$ is the function associated with the fitted curve of the CDM centerline in frame $\{cm_d\}$, $s^{'}$ denotes the arc length of the CDM centerline which its interval is from the origin of the frame $\{cm_p\}$ to the origin of the frame $\{cm_d\}$, and $L=35$ mm is the total arc length of the CDM centerline. Next, the curvature of each active area on the shape sensor centerline needs to be translated to the CDM centerline by:
\begin{equation}
\label{curvature-CDM-center-curve}
    \begin{aligned}
    \kappa_{cm}[j]=\frac{\kappa[j]}{1+d_{offset}\kappa[j]\text{cos}(\phi[j])},
    \end{aligned}
\end{equation}

where $\kappa_{cm}[j]$ is the curvature of the $j^{th}$ active area on the CDM centerline. Using \eqref{arc-length-in-P} and \eqref{curvature-CDM-center-curve}, the continuous curvature of the CDM center curve, $\kappa_{cm}(s^{'})$, can be determined by spline interpolation. Finally, considering the frame $\{{cm}_p\}$ as the frame $\{i\}$, and $s^{'}$ instead of $s$, the coordinates of the CDM centerline can be found using \eqref{theta-shape-sensor} and \eqref{curve of the shape sensor}.
\section{EXPERIMENTS AND RESULTS}
\subsection{Experimental Setup}
\begin{figure}
\centering
\includegraphics[width=\columnwidth]{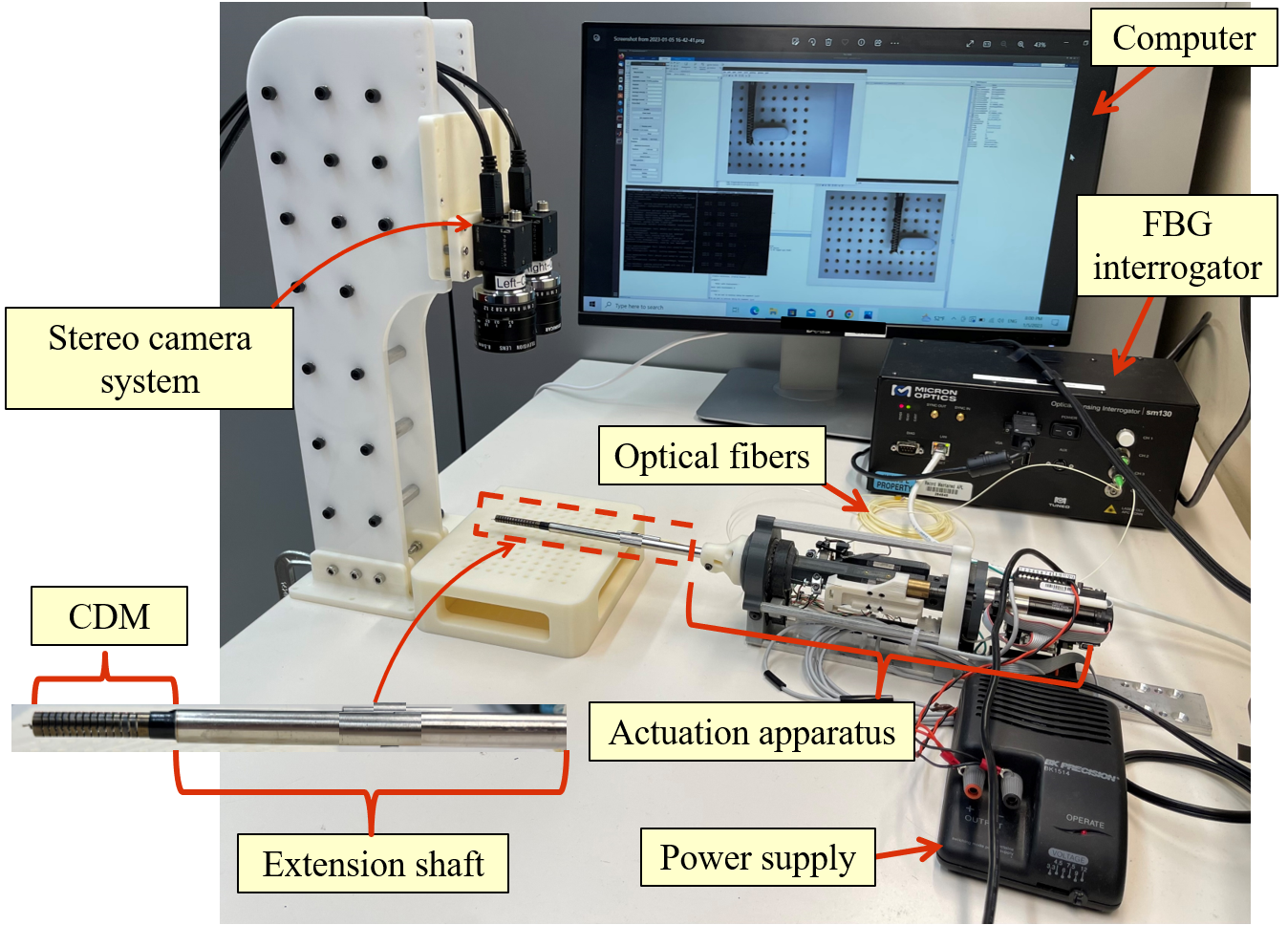}
\caption{The experimental setup, including the actuation apparatus, the stereo camera system, the FBG shape sensor.}
\label{Experimental-setup}
\end{figure}
The experimental setup included actuation apparatus, the shape sensing unit inside the CDM, and the stereo camera system ({\color{subsectioncolor}{Fig.\ref{Experimental-setup}}}). The actuation apparatus consisted of the linear actuator for driving the actuating cable of the CDM, and a holding block to keep DC motors and the CDM’s chunk. The linear actuator consisted of two graphite brushed DC-motors (4.5W, RE 16, Maxon, Switzerland) with a high-precision encoder (MR, Type M, 512 CPT, Maxon, Switzerland) and an integrated gear box-ball screw mechanism (Spindle Drive GP 16S, Maxon, Switzerland), as well as a position controller (EPOS2, Maxon, Switzerland) \cite{ma2021active}. While the CDM was bending, the FBG sensor wavelengths were measured by an optical sensing interrogator (sm130, Micron Optics Inc., Atlanta, GA) at a frequency of 100 Hz. Also, two FL2-08S2C cameras (Point Grey Research Inc., Richmond, BC, Canada), attached to the height-adjustable camera stand mount, captured the CDM images at a frequency of 30 Hz. To always make the CDM flexible segment visible to both cameras, the cameras were placed 30 cm above the CDM working area. The measurements for FBG wavelengths were performed using a C++ code, developed using the C++ CISST-SAW libraries\cite{kazanzides2014open}. The images of both stereo cameras and FBG wavelengths were recorded, and
%with a script written in MATLAB (MathWorks, Natick, MA, USA).  
equations \eqref{curvature}$-$\eqref{curvature-CDM-center-curve} were applied to reconstruct the CDM centerline using the measured wavelengths.
%Measured wavelengths were fed to MATLAB code for reconstructing the CDM centerline using \eqref{curvature}$-$\eqref{curvature-CDM-center-curve}.

\subsection{Image Processing}
\begin{figure}[!t]
\centering
\includegraphics[width=3.3in]{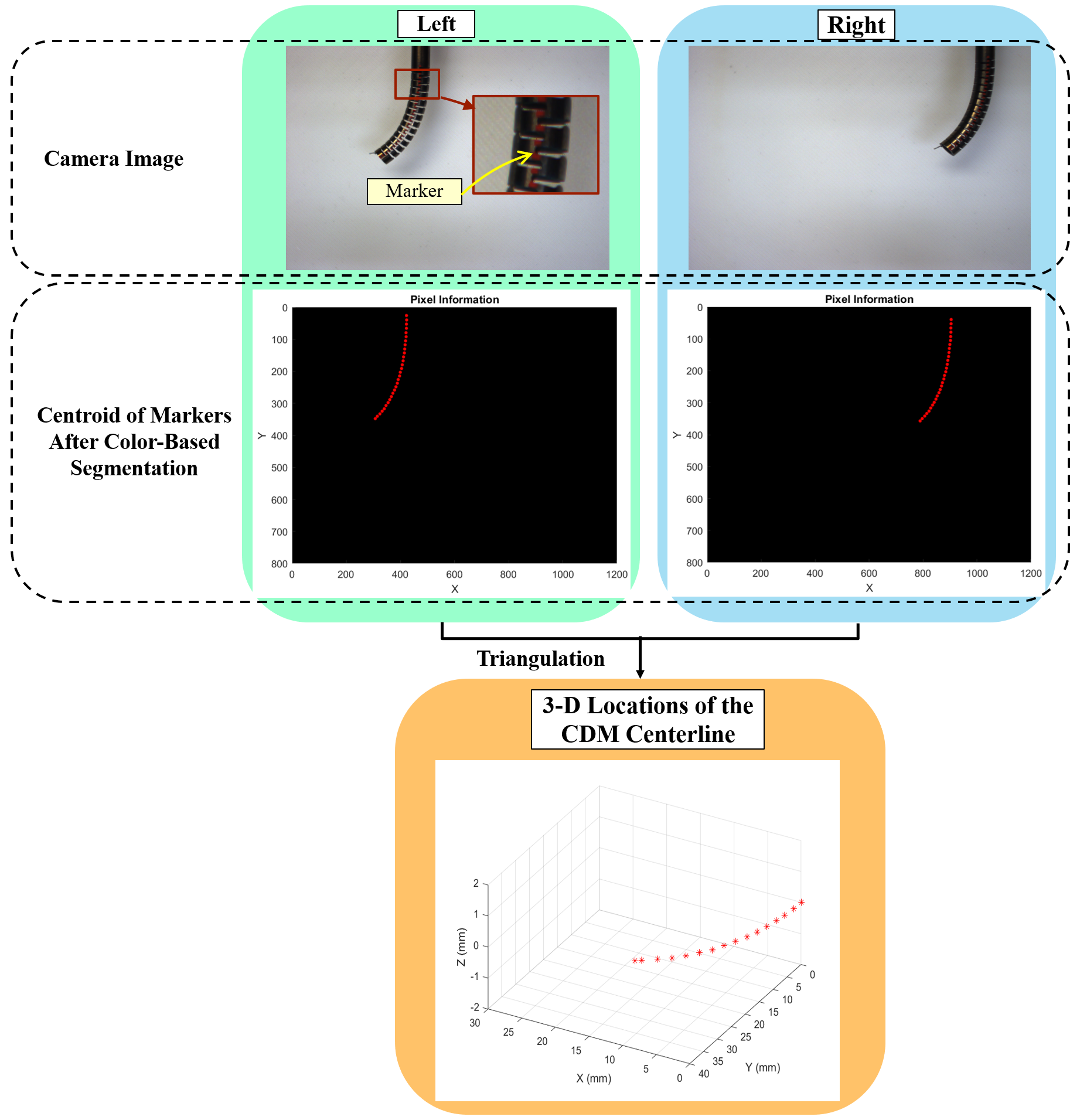}
\caption{Extraction of 3-D locations of the CDM centerline from stereo images using color-based segmentaion and triangulation.}
\label{Image-proccesing}
\end{figure}
A pair of stereo cameras with a resolution of 1024×768 was used to track red markers attached to the CDM centerline. To find the cameras' intrinsic and extrinsic matrices, the stereo camera pair was calibrated using the stereo camera calibration toolbox in MATLAB. The resulting overall mean error for the stereo pair calibration was 0.23 pixels. The measurement uncertainty of the stereo cameras after camera calibration was 0.253 mm, which was obtained by finding the Euclidean distance between two markers with predetermined positions on the calibration jig. The 3D locations of the centroids of the markers were calculated by developing a computer algorithm for color-based segmentation and triangulation of the images ({\color{subsectioncolor}{Fig.\ref{Image-proccesing}}}). The reconstructed CDM centerline from stereo images was then used as a ground truth to validate the CDM shape reconstruction model.  
%To find the 3-D locations of the centroid of the markers, an algorithm was written in MATLAB for color-based segmentation and triangulation of images ({\color{subsectioncolor}{Fig.\ref{Image-proccesing}}}). 

\subsection{Calibration}
\begin{figure}[!t]
\centering
\includegraphics[width=\columnwidth, height=2in]{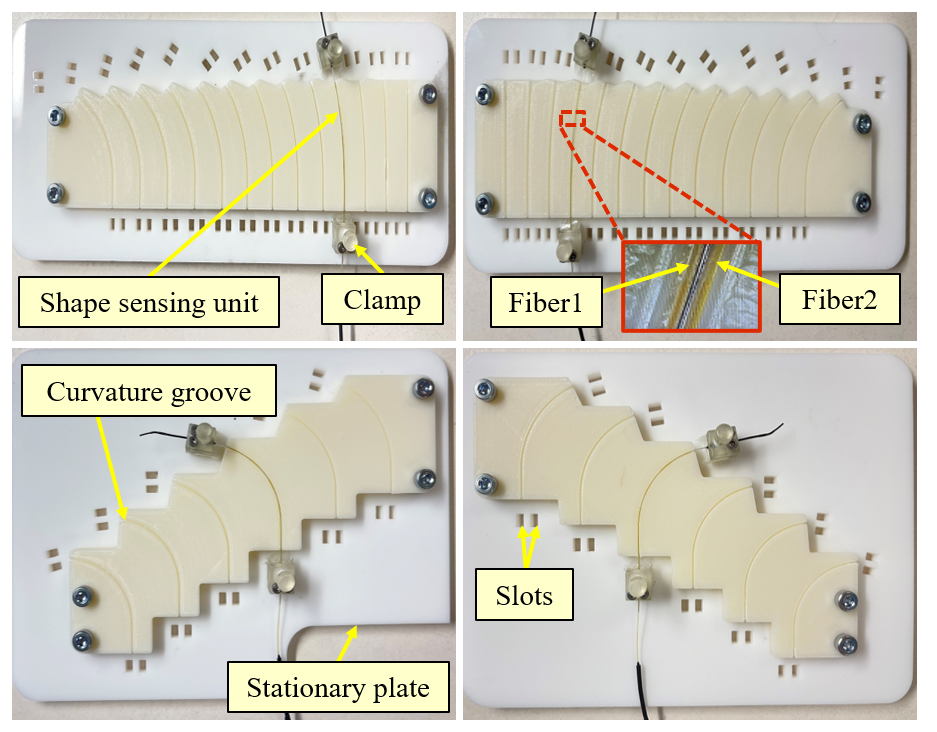}
\caption{Experimental setup for FBG sensors calibration using calibration jigs of positive and negative deflections with known curvatures.}
\label{FBG-Calibration-Jig}
\end{figure}

\begin{figure}
    \centering
    \subfigure[]
    {
        \includegraphics[width=\columnwidth, height=1.9in]{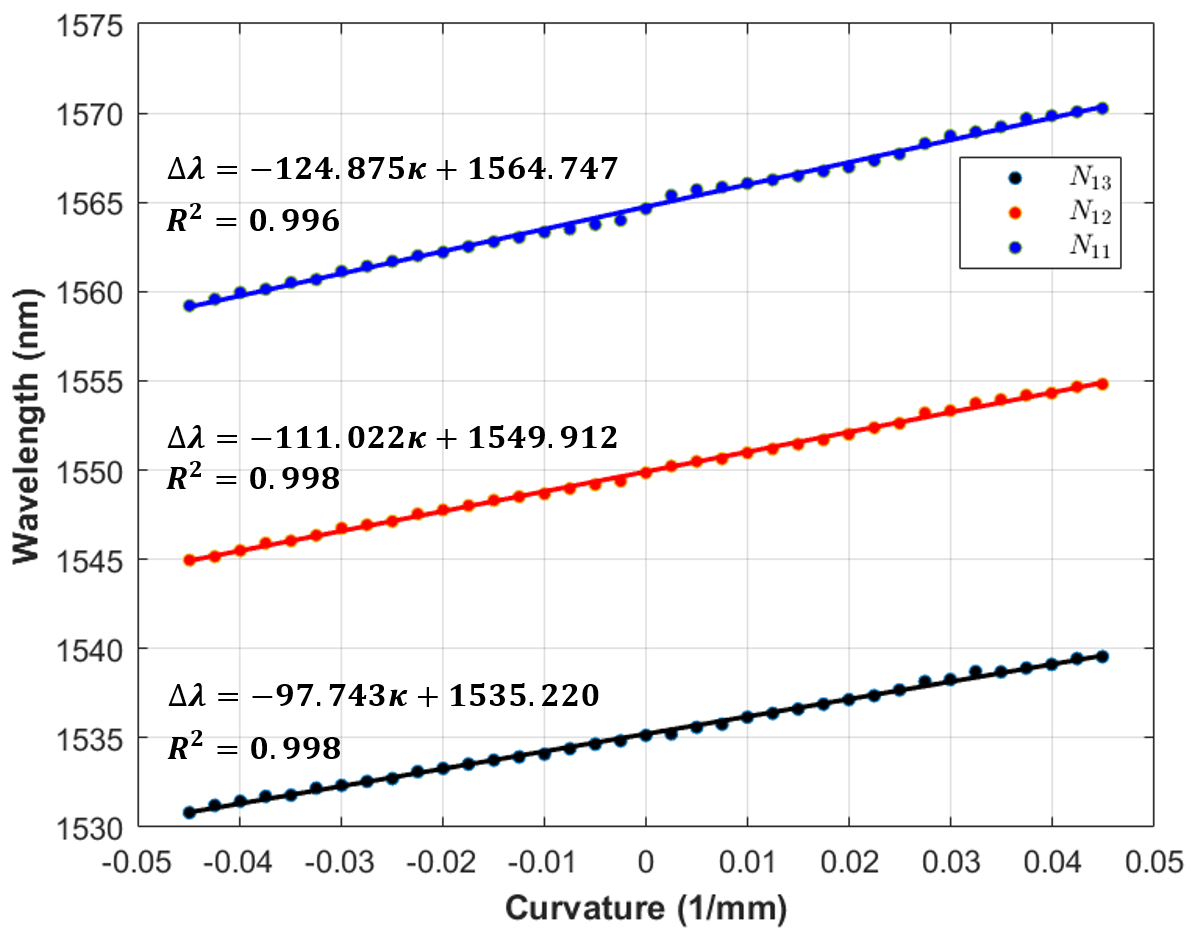}
        \label{fiber1_wave_curve}
    }
    \\
    \subfigure[]
    {
        \includegraphics[width=\columnwidth, height=1.9in]{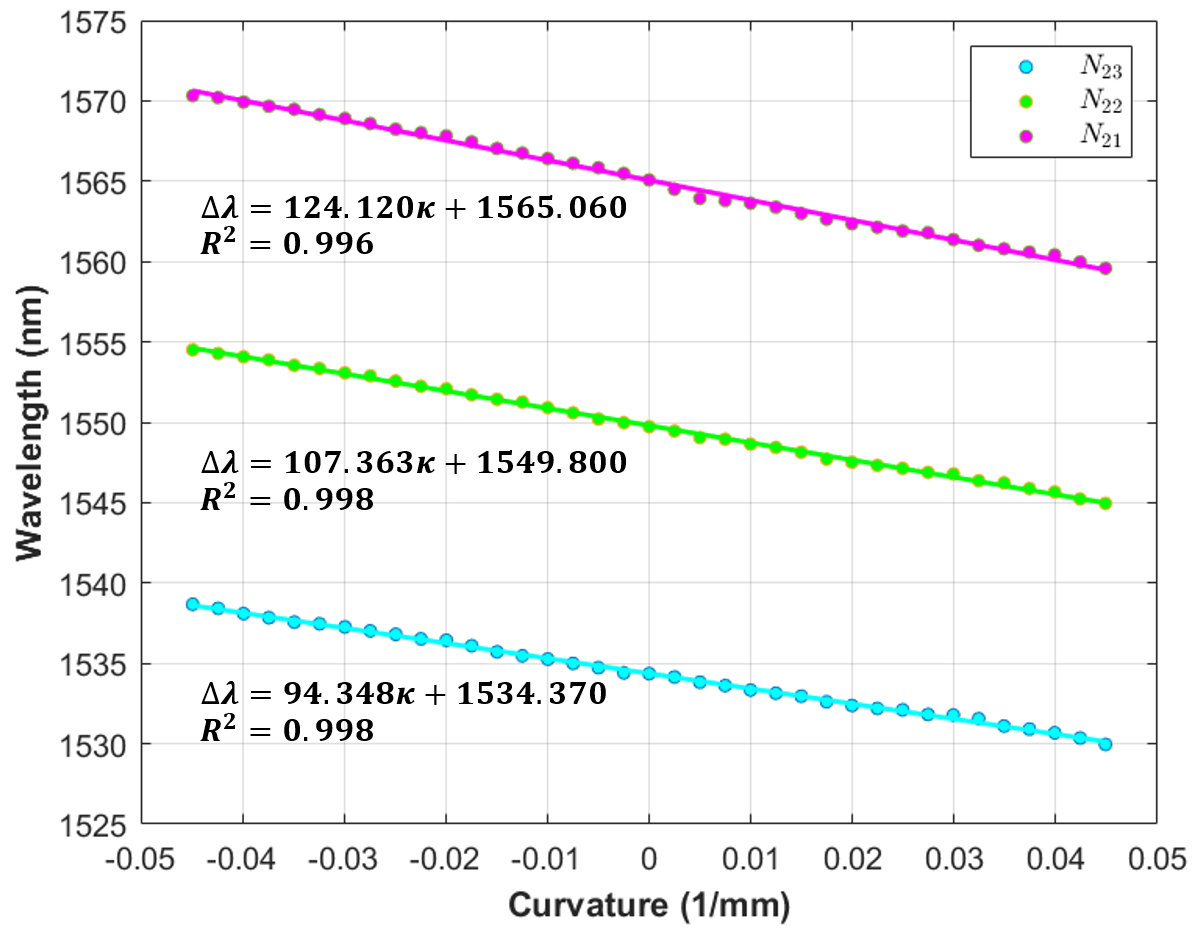}
        \label{fiber2_wave_curve}
    }
    \caption{Linear correlation between the Curvature and wavelength of the shape sensing unit at {\color{subsectioncolor}{(a)}} fiber1 and {\color{subsectioncolor}{(b)}} fiber2.}
    \label{FBGwave_curvature}
\end{figure}

To validate the linear relationship between the curvature and FBG wavelength, 3D-printed calibration jigs with discrete constant curvature grooves ranging from $-90^\circ$ to $90^\circ$ at $5^\circ$ interval and constant arc length were designed ({\color{subsectioncolor}{Fig.\ref{FBG-Calibration-Jig}}}). In this experiment, the sensing unit was examined at the optimal value of the sensor orthogonal distance, discussed in section II-C. The sensing unit was fixed between two clamps to maintain the sensing unit's orientation. %At each curvature groove, two clamps were inserted into the slots of the stationary plate to keep the sensing unit concentrically aligned with the inner wall of the curvature groove.%
The wavelengths of FBG nodes were sampled five times at each curvature groove, and then the mean value of data points was calculated. The sampling time was set to 3 seconds. {\color{subsectioncolor}{Fig.\ref{FBGwave_curvature}}} shows the calibration results of three FBG nodes at each fiber. The wavelength-curvature data points were well fitted with lines, with R squared ($R^2$) in the range of 0.996-0.998.

%\begin{figure}[!t]
%\centering
%\subfloat[]{\includegraphics[width=3in]{wave_curvature_Fiber1_ex.png}%
%\label{fiber1_wave_curve}}
%\hfil
%\subfloat[]{\includegraphics[width=3in]{wave_curvature_Fiber2_ex.png}%
%\label{fiber2_wave_curve}}
%\caption{Linear correlation between the Curvature and wavelength of the shape sensing unit at {\color{subsectioncolor}{(a)}} fiber1 and {\color{subsectioncolor}{(b)}} fiber2.}
%\label{FBGwave_curvature}
%\end{figure}
Due to the manual fabrication of the sensing unit and the limited accuracy in the building process, the exact values of the position, $r_{kj}$, and orientation, $\theta_{kj}$, of each FBG node are unknown. To find these unknown parameters, the data points of the wavelength-curvature experiment of the shape sensor for curvature grooves ranging from $-90^\circ$ to  $90^\circ$ at $10^\circ$ interval were used. Considering the pure bending at a constant temperature, the calibration process at the $j^{th}$ active area was accomplished by a least square optimization problem:
\begin{equation}
\label{sensor-least-square}
    \begin{aligned}
    \min_{P_j}\sum_{m=1}^N re_{m_j}^2 =\min_{P_j} \| y_{model_{j}}-y_{gt_j} \|_2^2,
    \end{aligned}
\end{equation}

where $re_{m_{j}}$ is the residual error of the $m^{th}$ observation, and $P_j$ is the set of unknown parameters for the positions and orientation of FBG nodes. $y_{gt_j}\in R^{N\times2}$ is a stack of \textit{N} ground truth data points for the curvature and bending direction, $y_{model_j}\in R^{N\times2}$ is a stack of \textit{N} observation data points for the curvature and bending direction, obtained by \eqref{curvature} and \eqref{curvature direction}. This minimization was implemented using least squares method. Results are shown in {\color{subsectioncolor}{Table \ref{geometric-params}}}. 
\begin{table}[t]
\caption{Calibrated ParameterS for Positions ($r_{K1}$, $r_{K2}$, $r_{K3}$) and Orientations ($\theta_{K1}$, $\theta_{K2}$, $\theta_{K3}$) of FBG Nodes on Each Fiber (\textit{K}=1,2)\label{geometric-params}}
\centering
\begin{tabular}{|c|c|c|c|c|c|c|}
\hline
\multirow{2}{*}{\makecell{Fiber index \\(\textit{k})}} & 
    \multicolumn{3}{|c|}{Position (\textit{mm})} &
    \multicolumn{3}{|c|}{Orientation (degree)} \\  \cline{2-7}
    & $r_{k1}$ & $r_{k2}$ & $r_{k3}$ & $\theta_{k1}$ & $\theta_{k2}$ & $\theta_{k3}$ \\
\hline
1 & 0.159 & 0.150 & 0.155 & 60.848 & 60.790 & 60.733 \\
\hline
2 & 0.159 & 0.158 & 0.154 & 60.848 & 60.790 & 60.733 \\
\hline
\end{tabular}
\end{table}
To validate the accuracy of the obtained positions and orientations of FBG nodes, the data points of the wavelength-curvature experiment of the shape sensor for curvature grooves ranging from $-85^\circ$ to $85^\circ$ at $10^\circ$ interval were used. Results demonstrate that the model of the sensing unit with equivalent position and orientation of each FBG node can predict the curvature and bending direction with the mean error of 4.704$e^{-4}$\textpm5.730$e^{-4}$ $mm^{-1}$ and 0.057\textpm0.019 degree, respectively.

\begin{figure}[!t]
\centering
\includegraphics[width=\columnwidth, height=1.9in]{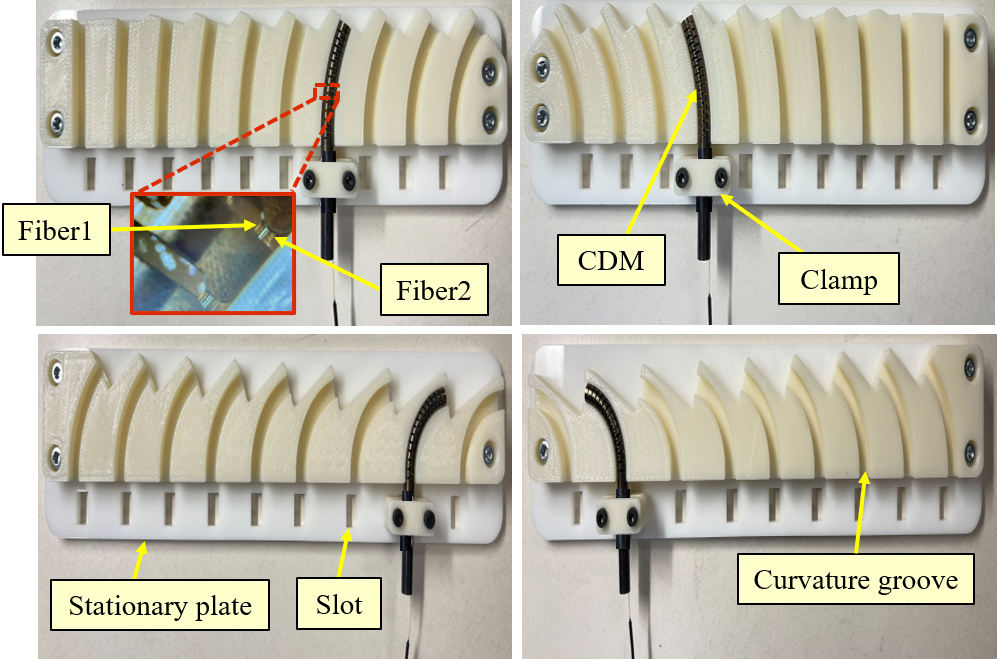}
\caption{Experimental setup for FBG sensors calibration inside the CDM using calibration jigs of positive and negative deflections with known curvatures.}
\label{CDM-Calibration-Jig}
\end{figure}

 Based on the deflection direction as well as the inconsistent effect of the hindered friction at interactions between the sensing unit and the CDM sensor channel \cite{gao2017general}, the wavelength shifts at each active area for positive and negative deflections are exactly the same. As discussed in section II-B, two sets of calibration coefficients are defined at each active area; $C_{p_{j}}$ and $C_{n_{j}}$ are the calibration coefficients of $j^{th}$ active area for positive and negative deflections, respectively. 3D-printed calibration jigs with discrete curvature grooves ranging from $-90^\circ$ to $90^\circ$ at $5^\circ$ interval were designed ({\color{subsectioncolor}{Fig.\ref{CDM-Calibration-Jig}}}). The curvature and total arc length of each groove are constant. The CDM was fixed at its proximal end by two clamps, which inserted into the slots of the stationary plate for maintaining the CDM’s bending plane parallel to the surface of the stationary plate. Five trials, one each with 3 seconds sampling time, were conducted at each groove. The wavelengths of FBG nodes from all five trials were averaged. The calibration process was performed by a least square optimization problem to find the set of coefficients that minimizes the curvature error at each active area. Experimental results of the CDM with embedded sensing unit for curvature grooves ranging from $-90^{\circ}$ to $90^{\circ}$ at $10^{\circ}$ interval were used for the least square problem which is given by:
\begin{equation}
\label{curvature-least-square}
    \begin{aligned}
    \min_{C_j}\sum_{m=1}^N re_{m_j}^2 =\min_{C_j} \| \kappa_{model_{j}}-\kappa_{gt_j} \|_2^2,
    \end{aligned}
\end{equation}

where $C_j$ is the unknown coefficient of either $C_{p_{j}}$ or $C_{n_{j}}$, based on the bending direction at the $j^{th}$ active area. $\kappa_{gt_j}\in R^{N}$ is a stack of \textit{N} ground truth data points for the curvature, $\kappa_{{model}_j}\in R^{N}$ is a stack of \textit{N} observation data points for the curvature, obtained by \eqref{curvature} and \eqref{curvature correction}. Results are shown in {\color{subsectioncolor}{Table \ref{calibration-coeff}}}.
\begin{table}[t]
\caption{Calibration Coefficients for Positive ($C_{p_{1}}$, $C_{p_{2}}$, $C_{p_{3}}$) and Negative ($C_{n_{1}}$, $C_{n_{2}}$, $C_{n_{3}}$) Deflections of the Sensing Unit Inside the CDM (\textit{j}=1,2,3)\label{calibration-coeff}}
\centering
\begin{tabular}{ |M{2.2cm}|M{2.2cm}|M{2.2cm}| }
\hline
\multirow{2}{*}{\makecell{Active area index \\(\textit{j})}} & 
    \multicolumn{2}{c|}{Calibration coefficient } \\ \cline{2-3}
    & $C_{p_{j}}$ & $C_{n_{j}}$ \\
\hline
1 & 1.024 & 0.917   \\
\hline
2 & 0.945 & 0.836  \\
\hline
3 & 0.985 & 0.655  \\
\hline
\end{tabular}
\end{table}
To find the accuracy of the obtained coefficients for curvature detection, the wavelength-curvature experimental results of the sensing unit inside the CDM for curvature grooves ranging from $-85^\circ$ to $85^\circ$ at $10^\circ$ interval were used. Results indicated that the model with the equivalent calibration coefficients can determine the curvature of positive and negative deflections with the mean error of 4.295$e^{-4}$\textpm4.585$e^{-4}$ $mm^{-1}$ and 4.414$e^{-4}$\textpm3.842$e^{-4}$ $mm^{-1}$, respectively. As discussed in section III-B, the sign of the wavelength shifts at each active area can be used to specify the corresponding calibration coefficient of the model. 
%if the sign of the wavelength shifts for fiber 1 and fiber 2 at the $j^{th}$ active area are positive and negative, respectively, $C_{p_{j}}$ will be chosen, otherwise, CDM is in negative deflection, and $C_{n_{j}}$ will be chosen.
\subsection{Static Experiment Conducted in Free Environment}
Two sets of experiments in a free environment including positive and negative deflections were conducted. Each set of experiments consists of two cycles: bending, and straightening. The bending cycle started from zero deflection at which the CDM was straight. Then, for each direction of deflection, the corresponding actuating cable was driven by the linear actuator, in the range of 0$-$5 mm, at 1 mm increments. When the actuating cable reached the maximum displacement, the straightening cycle was initiated such that the cable was released at 1 mm increments to return the CDM to its reference position. At each increment, the FBG wavelengths as well as the camera images of the CDM were recorded. Each set of experiments contains data points for 9 cable displacements, and repeated five times to assess the repeatability of the results. 

\begin{figure}[t]
    \centering
    \subfigure[]
    {
        \includegraphics[width=\columnwidth, height=2in]{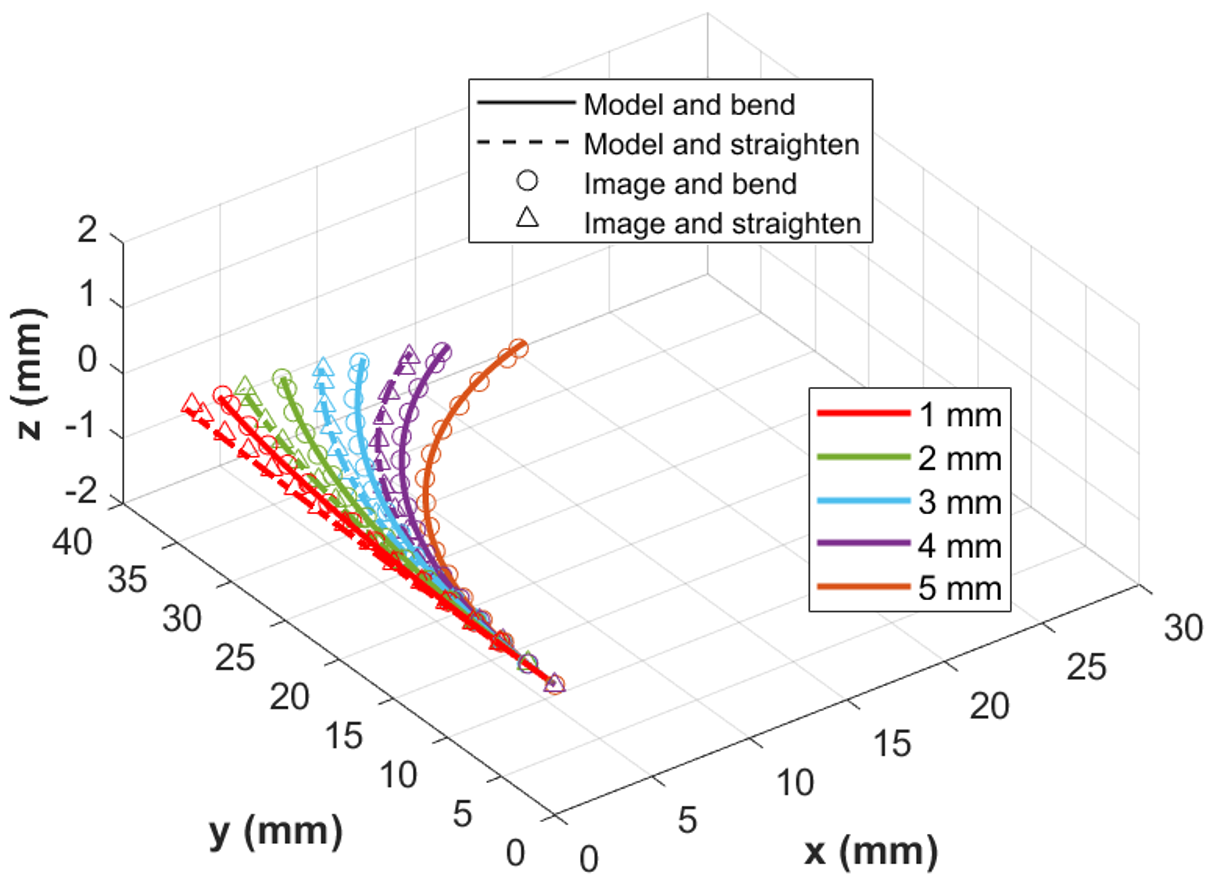}
        \label{CDM_centerline_posDef}
    }
    \\
    \subfigure[]
    {
        \includegraphics[width=\columnwidth, height=2in]{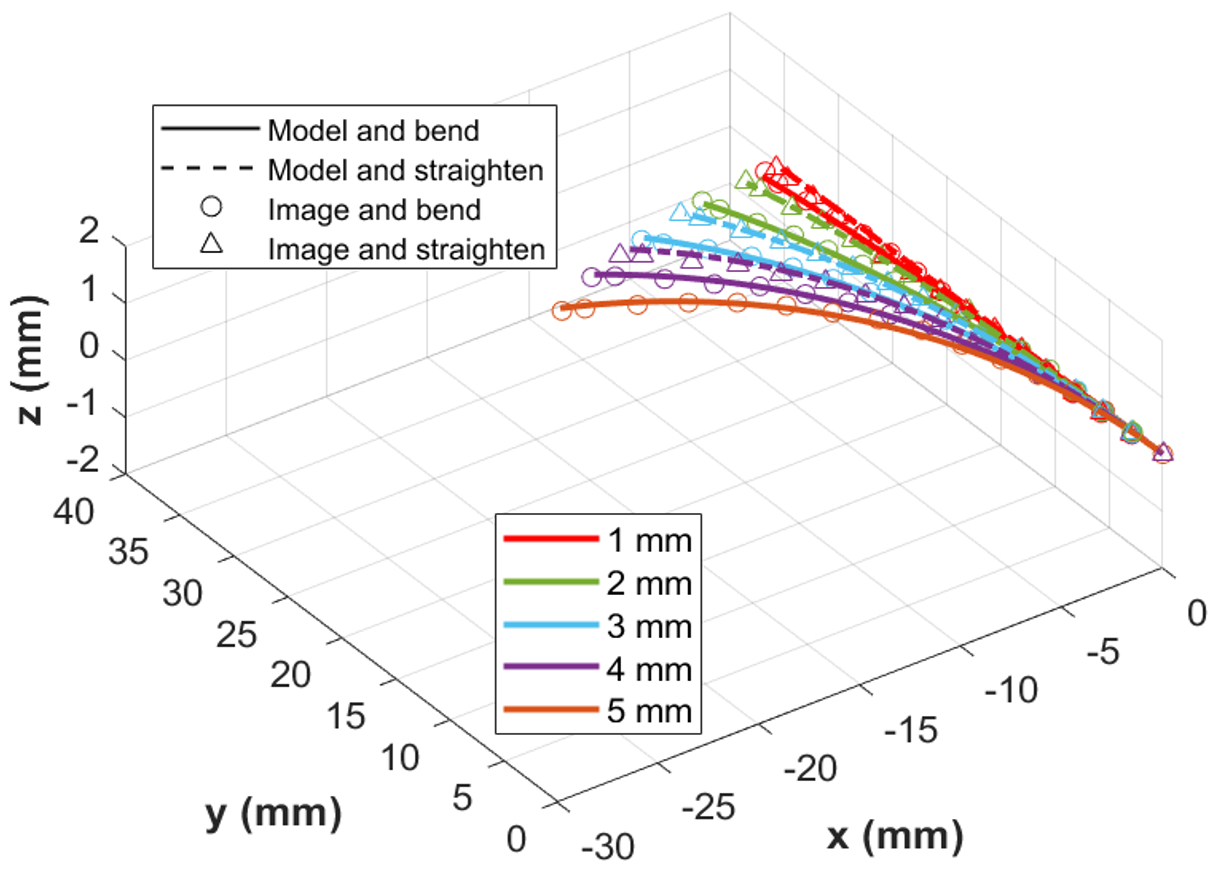}
        \label{CDM_centerline_negDef}
    }
    \caption{Shape reconstruction of the CDM in the free environment for {\color{subsectioncolor}{(a)}} positive deflection and {\color{subsectioncolor}{(b)}} negative deflection.}
    \label{3D_CDM_Centerline_PosNeg}
\end{figure}

The CDM's centerlines obtained from the reconstruction model and stereo images are illustrated in {\color{subsectioncolor}{Fig.\ref{3D_CDM_Centerline_PosNeg}}}. Each curve represents the average of five sampling data points. To compare the CDM centerlines with ground truth data, the mean absolute error and standard deviation of the CDM centerline and tip pose in addition to the maximum absolute error of the CDM centerline were calculated at each cable displacement. The errors are presented in {\color{subsectioncolor}{Table \ref{free-bending-error}}}. Results indicated a good agreement between the CDM centerline of the model and image in bending and straightening cycles with the overall mean error of 0.175\textpm0.081 mm for positive deflection, 0.262\textpm0.144 mm for negative deflection, and 0.216\textpm0.126 mm for positive/negative deflections. There was also a good agreement between the CDM tip pose of the model and image in bending and straightening cycles with the overall mean error of 0.197\textpm0.094 mm for positive deflection, 0.339\textpm0.202 mm for negative deflection, and 0.268\textpm0.172 mm for positive/negative deflections.
\begin{table*}
\caption{Shape Reconstruction and Tip Position Errors of the CDM in the Free Environment\label{free-bending-error}}
\centering
\begin{tabular}{ |c||c||c|c|c|c|c||c|c|c|c|c| }
\hline
& \multirow{3}{*}{\makecell{Cable displacement \\(\textit{mm})}} & 
    \multicolumn{5}{|c||}{Positive deflection } &
    \multicolumn{5}{|c|}{Negative deflection }\\ \cline{3-12}
    Cycle & & 
    \multicolumn{3}{|c|}{Shape error } & 
    \multicolumn{2}{|c||}{Tip error } &
    \multicolumn{3}{|c|}{Shape error } & 
    \multicolumn{2}{|c|}{Tip error }\\
    \cline{3-12}
    & & 
    \multicolumn{1}{|c|}{\makecell{Mean \\(\textit{mm})}} & 
    \multicolumn{1}{|c|}{\makecell{Std. \\(\textit{mm})}} &
    \multicolumn{1}{|c|}{\makecell{Max \\(\textit{mm})}} &
    \multicolumn{1}{|c|}{\makecell{Mean \\(\textit{mm})}} &
    \multicolumn{1}{|c||}{\makecell{Std. \\(\textit{mm})}} &
    
    \multicolumn{1}{|c|}{\makecell{Mean \\(\textit{mm})}} & 
    \multicolumn{1}{|c|}{\makecell{Std. \\(\textit{mm})}} &
    \multicolumn{1}{|c|}{\makecell{Max \\(\textit{mm})}} &
    \multicolumn{1}{|c|}{\makecell{Mean \\(\textit{mm})}} &
    \multicolumn{1}{|c|}{\makecell{Std. \\(\textit{mm})}} \\
    \hline
     & 1 & 
     0.126 & 0.065 & 0.261 & 
     0.204 & 0.003 &
     0.227 & 0.122 & 0.407 & 
     0.302 & 0.001 \\
    \cline{2-12}
     & 2 & 
     0.167 & 0.072 & 0.281 & 
     0.153 & 0.007 &
     0.218 & 0.129 & 0.467 & 
     0.064 & 0.006 \\
    \cline{2-12}
     Bending & 3 & 
     0.163 & 0.063 & 0.280 & 
     0.211 & 0.028 &
     0.252 & 0.118 & 0.488 & 
     0.201 & 0.014 \\
    \cline{2-12}
     & 4 & 
     0.166 & 0.066 & 0.283 & 
     0.250 & 0.017 &
     0.291 & 0.125 & 0.511 & 
     0.390 & 0.037 \\
    \cline{2-12}
     & 5 & 
     0.191 & 0.077 & 0.305 & 
     0.249 & 0.019 &
     0.357 & 0.159 & 0.617 & 
     0.559 & 0.110 \\
    \hline
     & 4 & 
     0.271 & 0.071 & 0.292 & 
     0.187 & 0.003 &
     0.296 & 0.190 & 0.656 & 
     0.656 & 0.004 \\
    \cline{2-12}
     & 3 & 
     0.139 & 0.064 & 0.276 & 
     0.188 & 0.009 &
     0.214 & 0.093 & 0.386 & 
     0.213 & 0.003 \\
    \cline{2-12}
     Straightening & 2 & 
     0.169 & 0.057 & 0.263 & 
     0.184 & 0.001 &
     0.204 & 0.104 & 0.377 & 
     0.170 & 0.004 \\
    \cline{2-12}
     & 1 & 
     0.142 & 0.008 & 0.267 & 
     0.185 & 0.004 &
     0.210 & 0.105 & 0.361 & 
     0.238 & 0.003 \\
\hline
\end{tabular}
\end{table*}

\subsection{Static Experiment Conducted in Presence of Obstacles}
\begin{figure}
\centering
\includegraphics[width=\columnwidth]{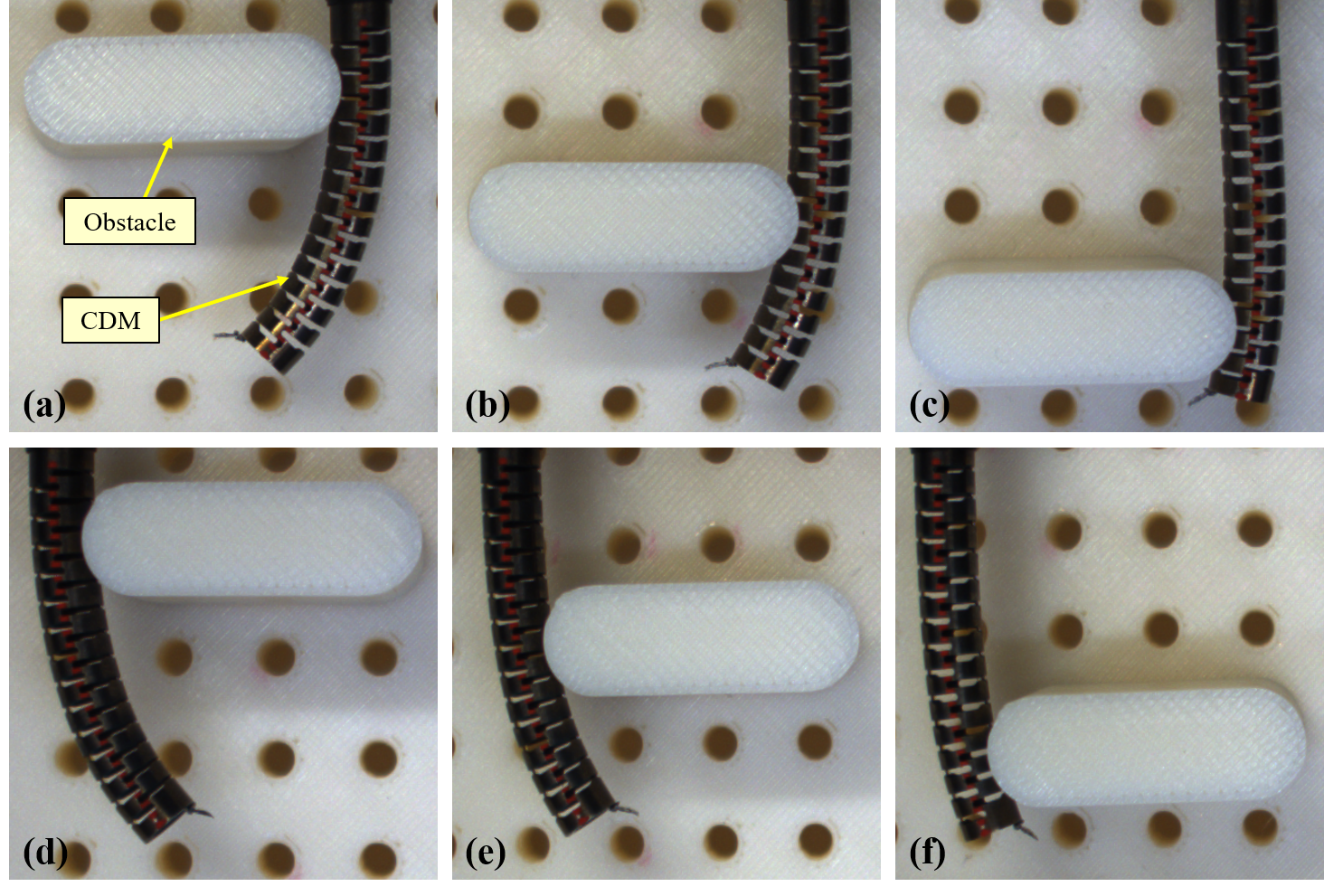}
\caption{Experimental setup of the CDM bending in constrained environment for {\color{subsectioncolor}{(a-c)}} positive deflection and {\color{subsectioncolor}{(d-f)}} negative deflection. Obstacles were placed at three different locations on each side of the CDM: {\color{subsectioncolor}{(a)}}, {\color{subsectioncolor}{(d)}} proximal; {\color{subsectioncolor}{(b)}}, {\color{subsectioncolor}{(e)}} middle; {\color{subsectioncolor}{(c)}}, {\color{subsectioncolor}{(f)}} distal.}
\label{Experimental-setup-constrained-environment}
\end{figure}

%In surgery, the CDM can perform various tasks in confined spaces, which involves interactions with soft tissues, lesions, and bone. 
To validate the shape reconstruction model in the case the contact forces applied along the CDM, two sets of experiments were conducted in the presence of obstacles for positive and negative deflections. As shown in {\color{subsectioncolor}{Fig.\ref{Experimental-setup-constrained-environment}}}, 3D-printed obstacles were placed at three different locations along the length of the CDM which was: 1) near the proximal end; 2) in the middle segment; 3) near the distal end. 

In each case, the segment of the CDM which interacted with the obstacle was obstructed from free bending and enforced the CDM to conform to more
complex shapes. Each set of experiments was repeated five times. {\color{subsectioncolor}{Fig.\ref{CDM_centerline_posDef_case1}}-(c)} and {\color{subsectioncolor}{Fig.\ref{CDM_centerline_negDef_case1}}-(f)} show the comparison between the ground truth data extracted from stereo images and the reconstruction model in the constrained environment for positive and negative deflections, respectively. The mean absolute error and standard deviation of the CDM centerline and tip as well as the maximum absolute error of the CDM centerline for three obstacle-interaction cases are given in {\color{subsectioncolor}{Table \ref{constrained-bending-error}}}. The overall mean tracking accuracy of the CDM centerline for positive/negative deflections was 0.436\textpm0.370 mm, 0.485\textpm0.418 mm, and 0.312\textpm0.261 mm, respectively, for proximal, middle, and distal cases of CDM bending with obstacles. There was also a good accuracy between the CDM tip position derived from the model and image in positive/negative deflections with the overall mean error of 0.811\textpm0.383 mm for the proximal case, 1.040\textpm0.327 mm for the middle case and 0.680\textpm0.422 mm for the distal case.

\section{Discussion}
This study presented a novel technique for building a thin, large deflection FBG-based shape-sensing unit that can be integrated into minimally invasive surgical systems. Our primary motivation is to develop a relatively inexpensive, easy-to-fabricate, and fast-fabricating (4 hours vs. days \cite{sefati2017highly,liu2015large}) shape sensing for a CDM that is designed for orthopaedic applications \cite{sefati2021dexterous,alambeigi2019use}. The CDM, therefore, is constrained to bend in one plane and provide maximum stability and resistance to bending in the plane orthogonal to the bending plane.  The proposed sensing method, however,  can be extended to the CDM designs with 3D bending. The shape reconstruction model was introduced to relate the wavelengths of FBG nodes to the curvature and bending direction of the CDM. The model was then implemented on the CDM and its efficacy in free and constrained environments for positive and negative deflections were assessed. 
\begin{figure*}
    \centering
    \subfigure[]
    {
        \includegraphics[width=2.2in]{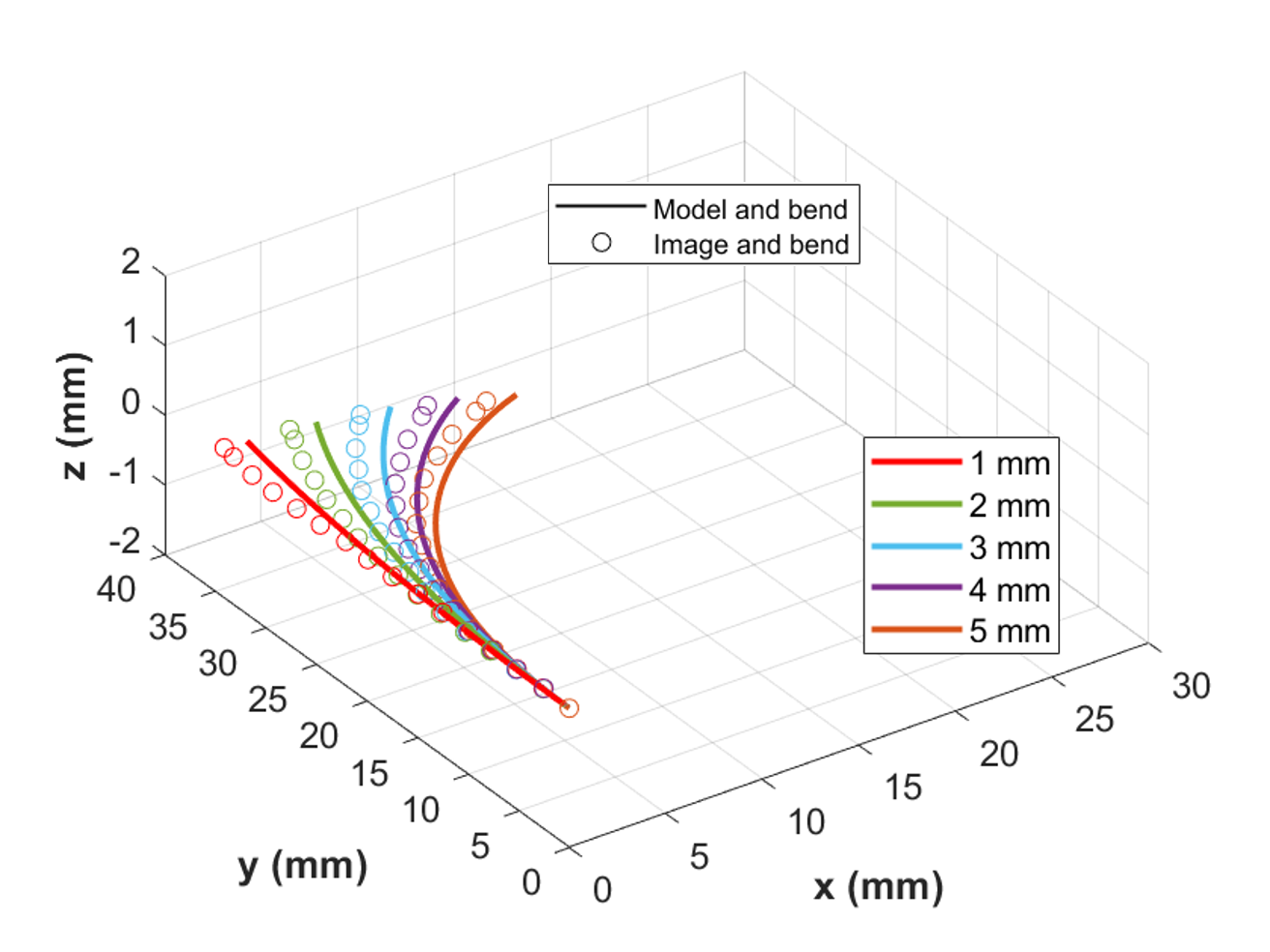}
        \label{CDM_centerline_posDef_case1}
    }
    \hfill
    \subfigure[]
    {
        \includegraphics[width=2.2in]{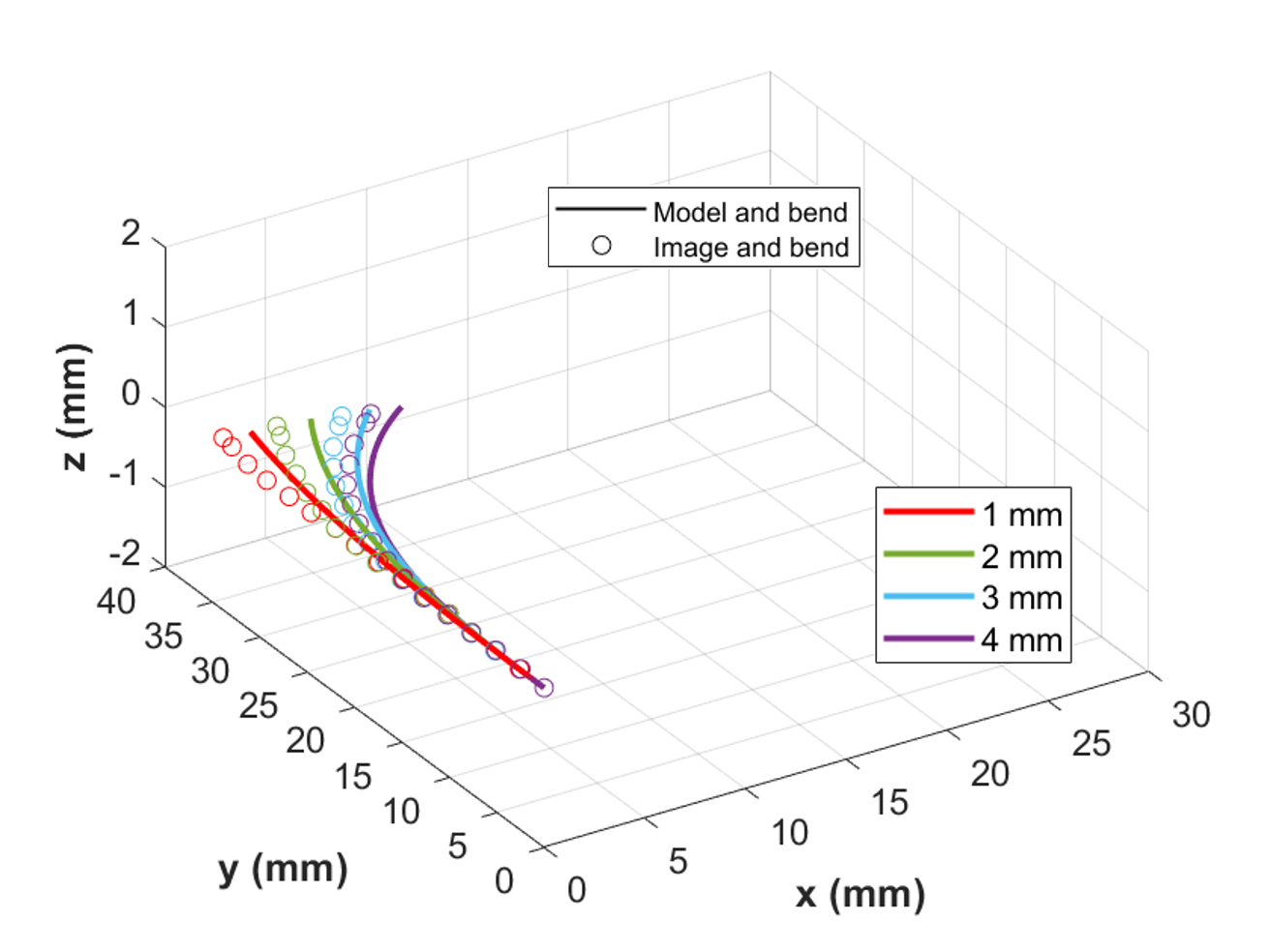}
        \label{CDM_centerline_posDef_case2}
    }
    \hfill
    \subfigure[]
    {
        \includegraphics[width=2.2in]{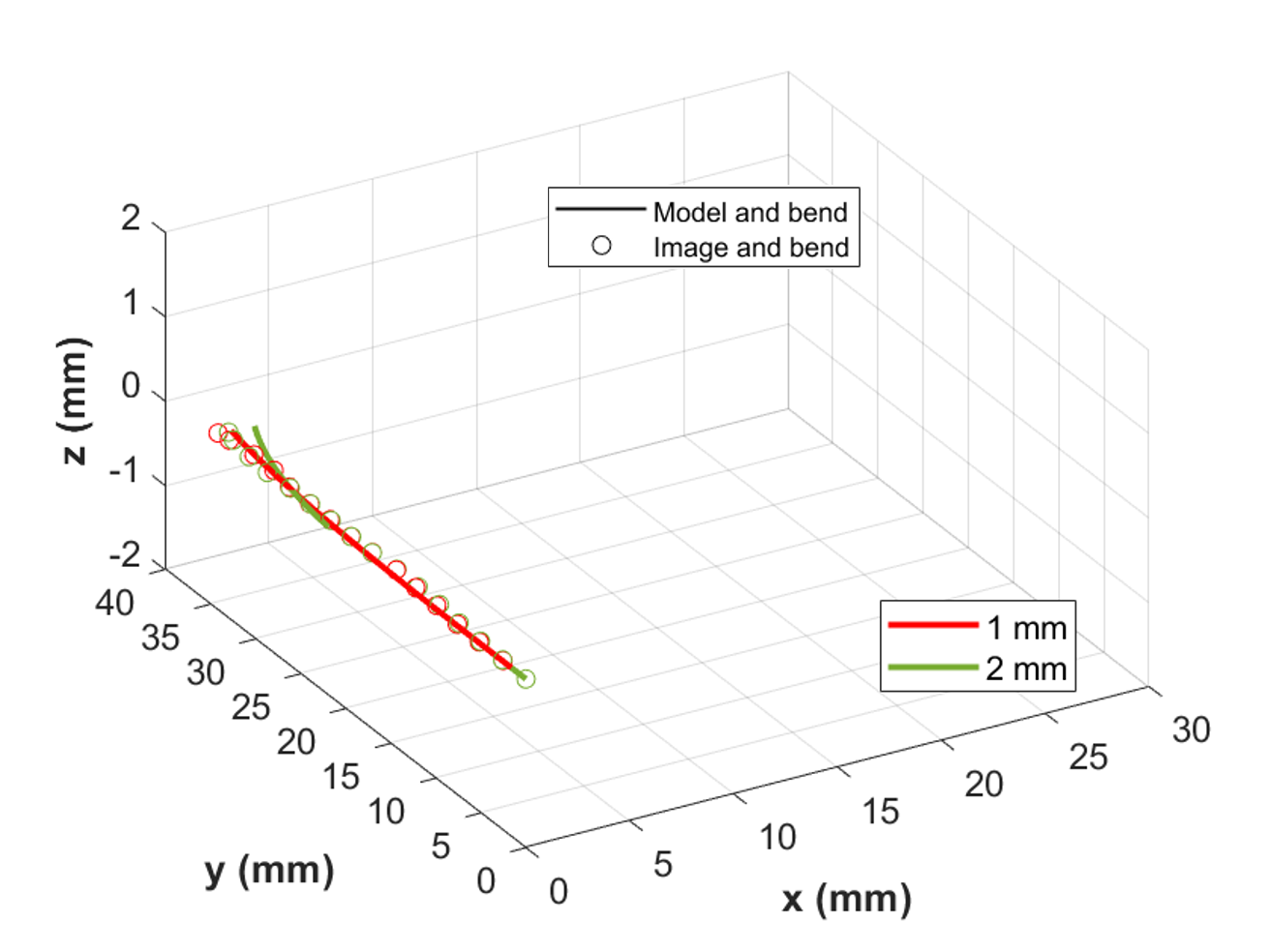}
        \label{CDM_centerline_posDef_case3}
    }
    \\
    \subfigure[]
    {
        \includegraphics[width=2.2in]{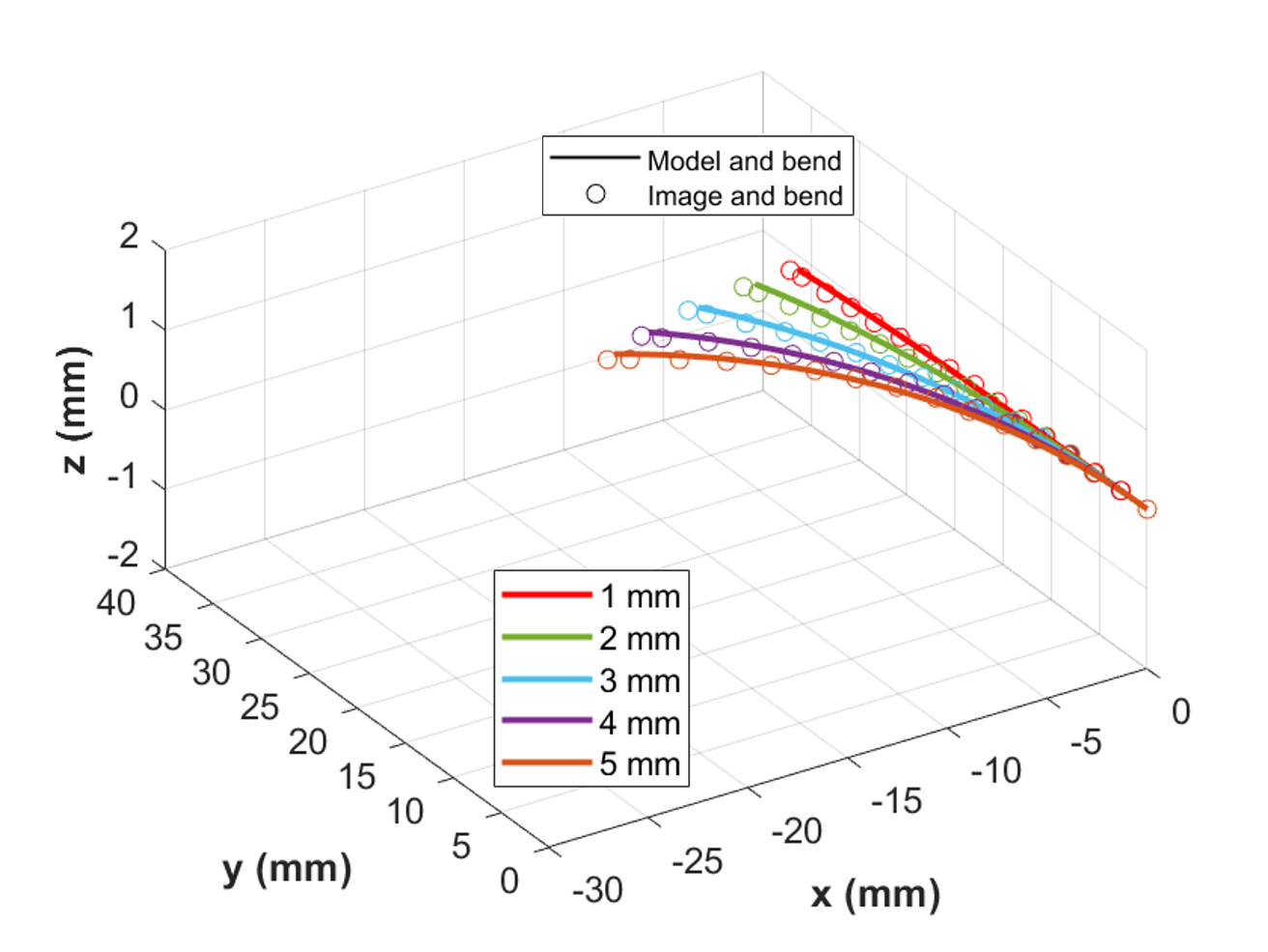}
        \label{CDM_centerline_negDef_case1}
    }
    \hfill
    \subfigure[]
    {
        \includegraphics[width=2.2in]{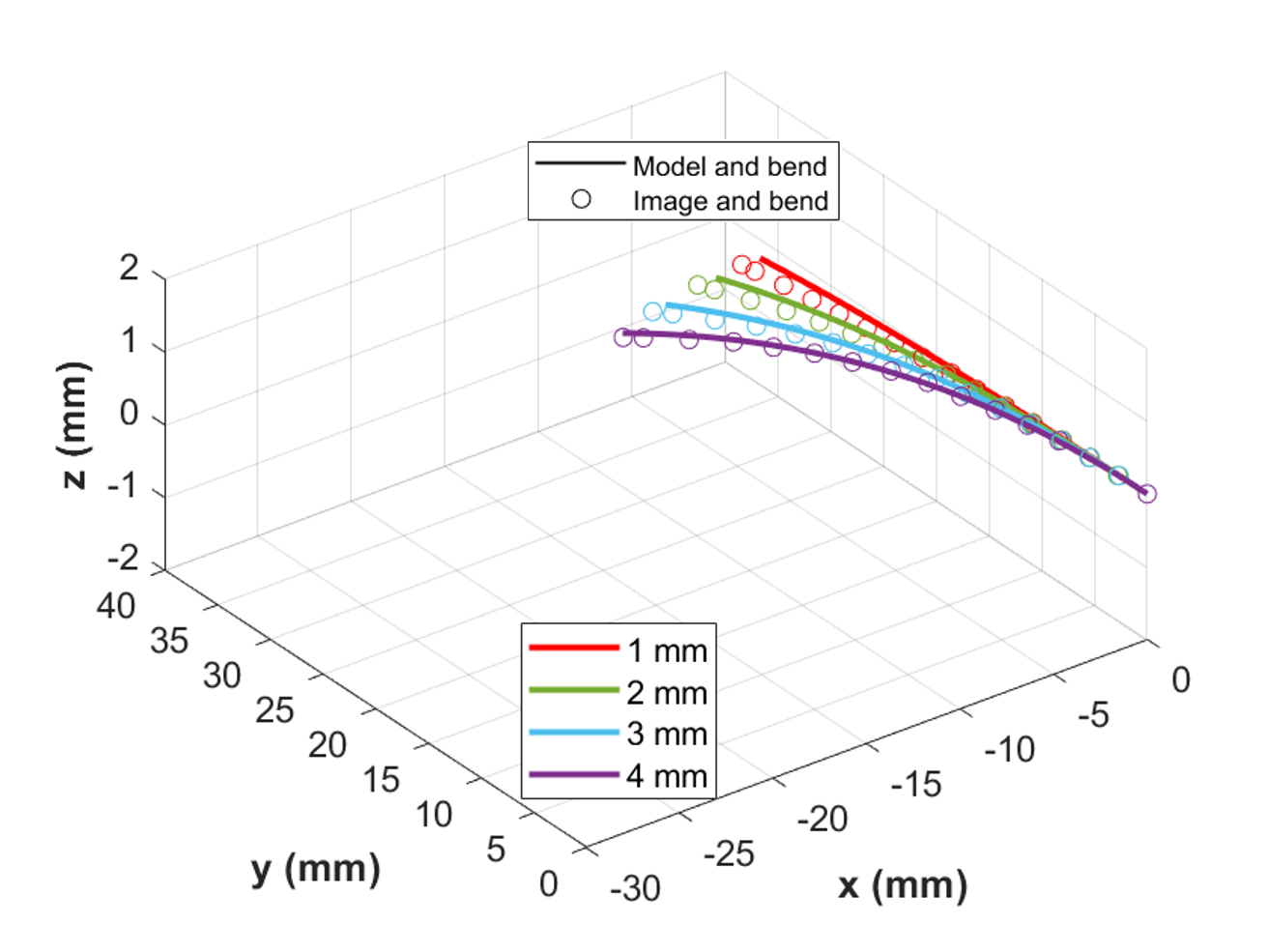}
        \label{CDM_centerline_negDef_case2}
    }
    \hfill
    \subfigure[]
    {
        \includegraphics[width=2.2in]{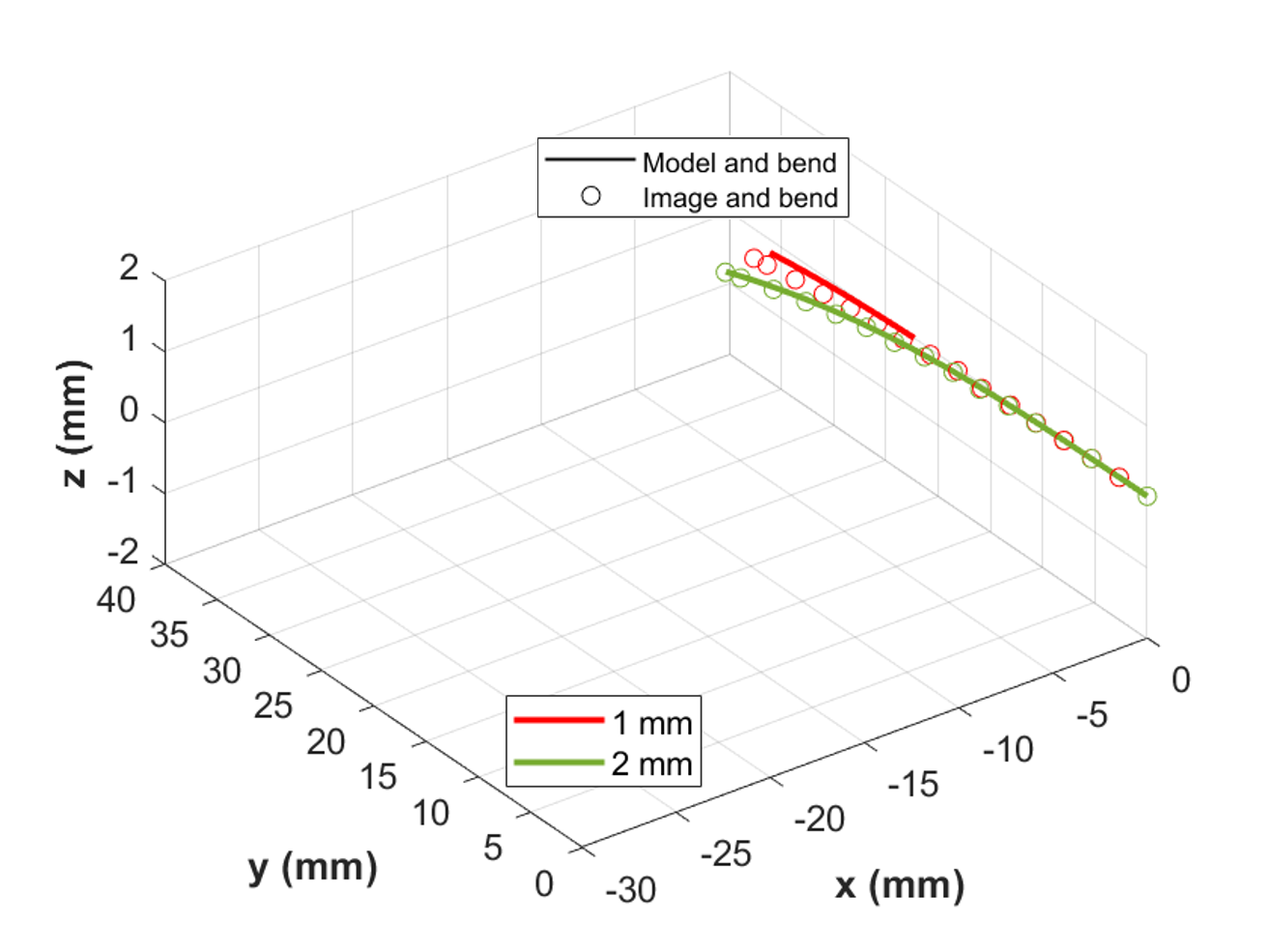}
        \label{CDM_centerline_negDef_case3}
    }
    \caption{Shape reconstruction of the shape sensing unit in the constrained environment for {\color{subsectioncolor}{(a}}-{\color{subsectioncolor}{c)}} positive deflection and {\color{subsectioncolor}{(d}}-{\color{subsectioncolor}{f)}} negative deflection in three different obstacle-interaction cases: {\color{subsectioncolor}{(a)}}, {\color{subsectioncolor}{(d)}} proximal; {\color{subsectioncolor}{(b)}}, {\color{subsectioncolor}{(e)}} middle; {\color{subsectioncolor}{(c)}}, {\color{subsectioncolor}{(f)}} distal.}
    \label{3D_CDM_Centerline_constrained_environment}
\end{figure*}
\begin{table*}[!t]
\caption{Shape reconstruction and tip position errors of the CDM in the constrained environment\label{constrained-bending-error}}
\centering
\begin{tabular}{ |M{1.8cm}||M{2.2cm}||c|c|c|c|c||c|c|c|c|c| }
\hline
\multirow{3}{*}{\makecell{Obstacle status}}
& \multirow{3}{*}{\makecell{Cable displacement \\(\textit{mm})}} & 
    \multicolumn{5}{|c||}{Positive deflection } &
    \multicolumn{5}{|c|}{Negative deflection }\\ \cline{3-12}
    & & 
    \multicolumn{3}{|c|}{Shape error } & 
    \multicolumn{2}{|c||}{Tip error } &
    \multicolumn{3}{|c|}{Shape error } & 
    \multicolumn{2}{|c|}{Tip error }\\
    \cline{3-12}
    & & 
    \multicolumn{1}{|c|}{\makecell{Mean \\(\textit{mm})}} & 
    \multicolumn{1}{|c|}{\makecell{Std. \\(\textit{mm})}} &
    \multicolumn{1}{|c|}{\makecell{Max \\(\textit{mm})}} &
    \multicolumn{1}{|c|}{\makecell{Mean \\(\textit{mm})}} &
    \multicolumn{1}{|c||}{\makecell{Std. \\(\textit{mm})}} &
    
    \multicolumn{1}{|c|}{\makecell{Mean \\(\textit{mm})}} & 
    \multicolumn{1}{|c|}{\makecell{Std. \\(\textit{mm})}} &
    \multicolumn{1}{|c|}{\makecell{Max \\(\textit{mm})}} &
    \multicolumn{1}{|c|}{\makecell{Mean \\(\textit{mm})}} &
    \multicolumn{1}{|c|}{\makecell{Std. \\(\textit{mm})}} \\
    \hline
    \multirow{5}{*}{\makecell{Proximal}}& 1 & 
     0.486 & 0.412 & 1.162 & 
     1.158 & 0.027 &
     0.171 & 0.114 & 0.362 & 
     0.360 & 0.011 \\
    \cline{2-12}
     & 2 & 
     0.682 & 0.415 & 1.298 & 
     1.267 & 0.023 &
     0.219 & 0.163 & 0.548 & 
     0.438 & 0.018 \\
    \cline{2-12}
     & 3 & 
     0.753 & 0.436 & 1.477 & 
     1.321 & 0.108 &
     0.210 & 0.121 & 0.539 & 
     0.478 & 0.038 \\
    \cline{2-12}
     & 4 & 
     0.700 & 0.375 & 1.360 & 
     1.117 & 0.168 &
     0.245 & 0.121 & 0.606 & 
     0.533 & 0.148 \\
    \cline{2-12}
     & 5 & 
     0.620 & 0.337 & 1.264 & 
     0.996 & 0.277 &
     0.276 & 0.161 & 0.812 & 
     0.547 & 0.217 \\

    \hline
     \multirow{4}{*}{\makecell{Middle}}& 1 & 
     0.512 & 0.466 & 1.323 & 
     1.315 & 0.055 &
     0.335 & 0.200 & 0.841 & 
     0.837 & 0.112 \\
    \cline{2-12}
     & 2 & 
     0.768 & 0.567 & 1.440 & 
     1.554 & 0.059 &
     0.387 & 0.220 & 0.716 & 
     0.896 & 0.114 \\
    \cline{2-12}
     & 3 & 
     0.675 & 0.401 & 1.256 & 
     1.157 & 0.074 &
     0.275 & 0.254 & 0.820 & 
     0.798 & 0.107 \\
    \cline{2-12}
     & 4 & 
     0.643 & 0.384 & 1.267 & 
     1.124 & 0.107 &
     0.283 & 0.187 & 0.790 & 
     0.635 & 0.120 \\
     
     \hline
     \multirow{2}{*}{\makecell{Distal}}& 1 & 
     0.235 & 0.144 & 0.617 & 
     0.614 & 0.012 &
     0.391 & 0.167 & 0.542 & 
     0.540 & 0.011 \\
    \cline{2-12}
     & 2 & 
     0.479 & 0.328 & 1.210 & 
     1.199 & 0.019 &
     0.242 & 0.175 & 0.274 & 
     0.366 & 0.021 \\
     
\hline
\end{tabular}
\end{table*}
The present technique for building the enclosed sensing unit was found to be easy, time-saving, repeatable, and cost-effective \cite{khan2019multi,moon2014fiber}. The linear wavelength-curvature relationship of the developed shape-sensing unit ({\color{subsectioncolor}{Fig.\ref{FBGwave_curvature}}}), had a pattern similar to previous studies \cite{liu2015large,sefati2017highly,sefati2016fbg,kim2017shape,song2021towards}. The sensing unit has a high sensitivity, up to 5.50 \nicefrac{nm}{mm}, at the $90^\circ$ bending angle compared to Sefati \cite{sefati2017highly}, up to 3 \nicefrac{nm}{mm}, and Liu \cite{liu2015large}, up to 4 \nicefrac{nm}{mm}. The sensitivity-enhancing property enables the sensing unit to detect small changes in curvature.

The results of the static experiments in the free environment ({\color{subsectioncolor}{Fig.\ref{3D_CDM_Centerline_PosNeg}}} and {\color{subsectioncolor}{Table \ref{free-bending-error}}}) and constrained environment ({\color{subsectioncolor}{Fig.\ref{3D_CDM_Centerline_constrained_environment}}} and {\color{subsectioncolor}{Table \ref{constrained-bending-error}}}) indicate that the reconstruction model for tracking the CDM centerline is comprehensive and effective and justifies the building approach of the FBG-based shape sensing unit employed in our study. The reconstruction model could track the CDM centerline, which is extracted from stereo images, with a maximum shape deviation of 0.305 mm for positive deflection and 0.656 mm for negative deflection in the free environment ({\color{subsectioncolor}{Table \ref{free-bending-error}}}), both below 2 percent of the CDM length, confirming the CDM is symmetric with respect to the deflection. The good consistency of the reconstruction model results in the constrained environment, implied by the maximum shape deviation of 1.477 mm for positive deflection and 0.841 mm for negative deflection ({\color{subsectioncolor}{Table \ref{constrained-bending-error}}}), suggests that the developed sensing unit and the reconstruction model are efficient for shape tracking of the CDM, eliminating the need for integrated multi-core optical fibers. Nevertheless, for the same actuating cable displacement, the mean errors of the CDM shape tracking in obstacle-interaction cases were slightly higher than free bending. This may be due to the clearance between the sensing unit and the CDM sensor channel. Considering the free movement of the sensing unit inside the CDM sensor channel with its tip fixed at the distal end and the frictional effects within the channel, the rigor of the model was higher than the models developed for the sensing unit directly attached through the length of the medical instruments \cite{khan2020pose,donder2021kalman,al2020improved}. Moreover, compared to the data-driven approach which estimated the CDM centerline by solving a constrained optimization problem associated with the C-shaped bending \cite{sefati2020data}, the proposed reconstruction model can be applied to S shape CDMs.

A limitation of the approach included the asymmetrical friction and local twist of the sensing unit affecting the shape reconstruction model by causing error in shape sensing. Although these errors due to the free motion of the sensing unit inside the CDM sensor channel were reduced by applying the calibration coefficients, the small clearance between the sensing unit and the CDM sensor channel is still a source of error for the curvature estimation. Local twist measurement and incorporating its effects remain for the future studies. 
%Future studies might employ more advanced shape reconstruction models to measure local twists for compensation. 
%The limited number of FBG nodes, also, leads to an inaccurate interpolation of the curvature and the bending direction of the CDM along its arc length. 
In addition, small errors in curvature estimation especially at the proximal end of the CDM can cause error accumulation through the CDM arclength. 
%Increasing the number of 
Depending on the accuracy requirements for a specified application, the number of FBG nodes at each shape sensing unit and/or equipping the CDM with two shape sensors may improve the shape reconstruction model. 

Hysteresis is another source of the error for CDM shape tracking ({\color{subsectioncolor}{Fig.\ref{3D_CDM_Centerline_PosNeg}}}). Hysteresis behavior in the CDM is due to the factors including friction between the actuating cable and the CDM cable channel, backlash of the actuating cables, and the hysteresis property of the NiTi. The effect of hysteresis may require additional modeling and must be considered in the process of designing feedback control paradigms for the CDM. 

In our study, the thermal conductivity of the NiTi rod used in the shape sensing unit and the CDM body reduces the temperature gradient through the sensor length, which can decrease the influence of the temperature on the wavelength shifts of FBG nodes. Although this assumption in general is compatible with FBG-based shape sensors \cite{cao2022spatial}, if needed, the shape reconstruction model can be further extended to account for changes in temperature. 
\section{Conclusion}
The goal of the study was to design, implement, and validate a novel shape
reconstruction model, based on a CDM equipped with a novel FBG-based sensing unit developed in-house. The fabrication process of the FBG shape senor was easy and fast, using a polycabonate tube with three lumens as a flexible enclosed substrate. The sensing model accounted for the internal twist compensation and the hindered friction compensation between the sensing unit and the CDM sensor channel. To reconstruct the CDM centerline, the wavelength measurements of the FBG nodes were utilized to determine the curvature and bending direction of the CDM. The results of experiments demonstrated the linear wavelength-curvature relationship with high sensitivity and large bending capability. Moreover, it was shown in static experiments
that the shape reconstruction model can accurately track the CDM centerline in free and constrained environments. It was concluded that the shape sensing of the CDM in the presence of non-constant curvature bending is feasible using the proposed sensing model.

\section*{Acknowledgment}
The author would like to thank Dr. S. Sefati for his contribution with the development of the software infrastructure.
%The author would like to thank Ms. A. Goodridge for her contribution with the development of the FBG shape sensor, and Mr. M. Esfandiari for his help with image processing.
\bibliographystyle{IEEEtran}
\bibliography{ref}

\newpage
\begin{IEEEbiography}[{\includegraphics[width=1in,height=1.25in,clip,keepaspectratio]{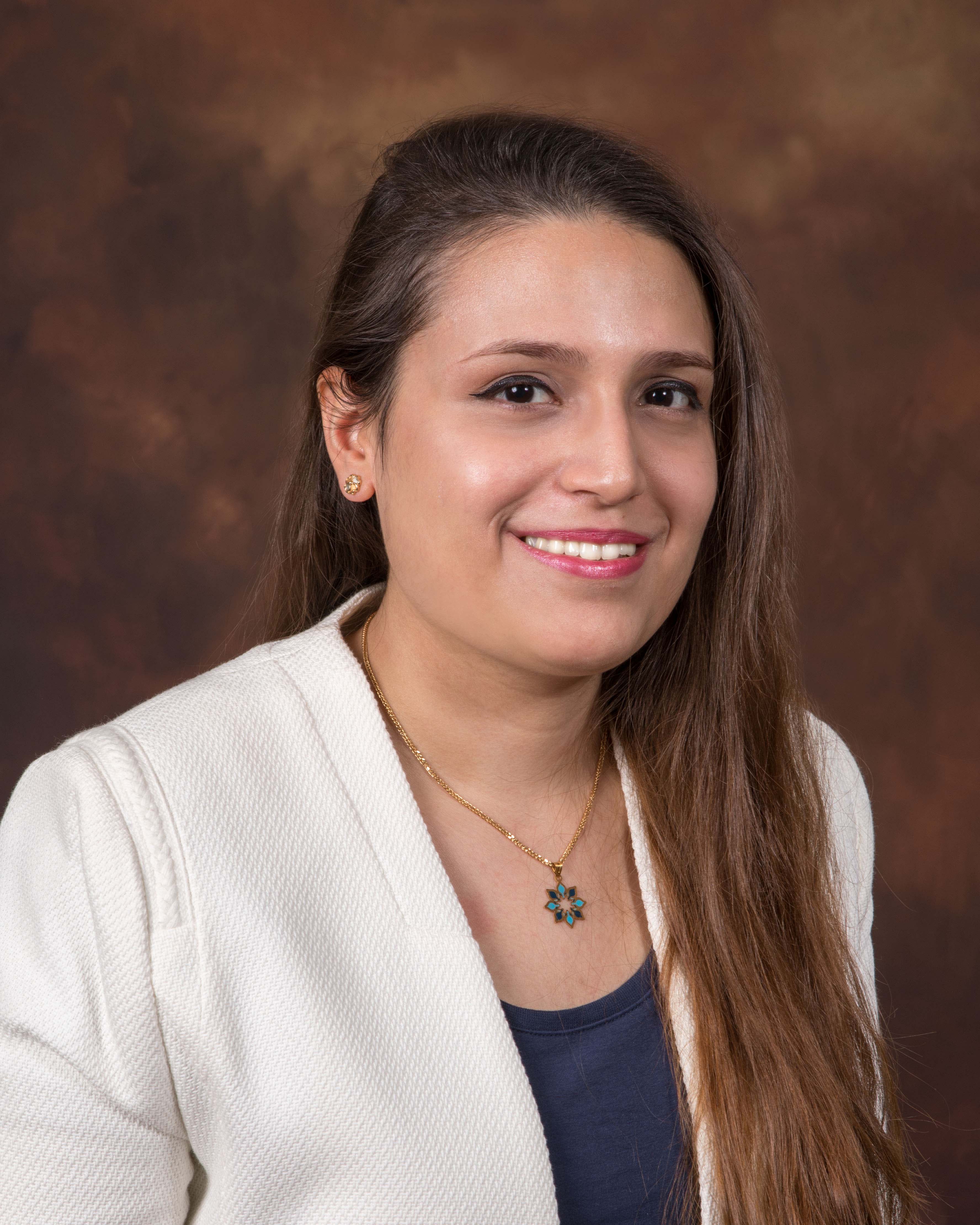}}]{Golchehr Amirkhani} received the B.Sc. degree in Mechanical Engineering from Shahid Beheshti University, Tehran, Iran, in 2015, and the M.Sc.
degrees in Mechatronics from Sharif University of
Technology, Tehran, Iran, in 2018. She is currently working toward the Ph.D. degree in Mechanical Engineering with Johns Hopkins University, Baltimore, MD, USA.

Her research interests include sensing and control of continuum manipulators with applications
in surgical robotics and autonomous systems.
\end{IEEEbiography}

\vspace{-1cm}
\begin{IEEEbiography}[{\includegraphics[width=1in,height=1.25in,clip,keepaspectratio]{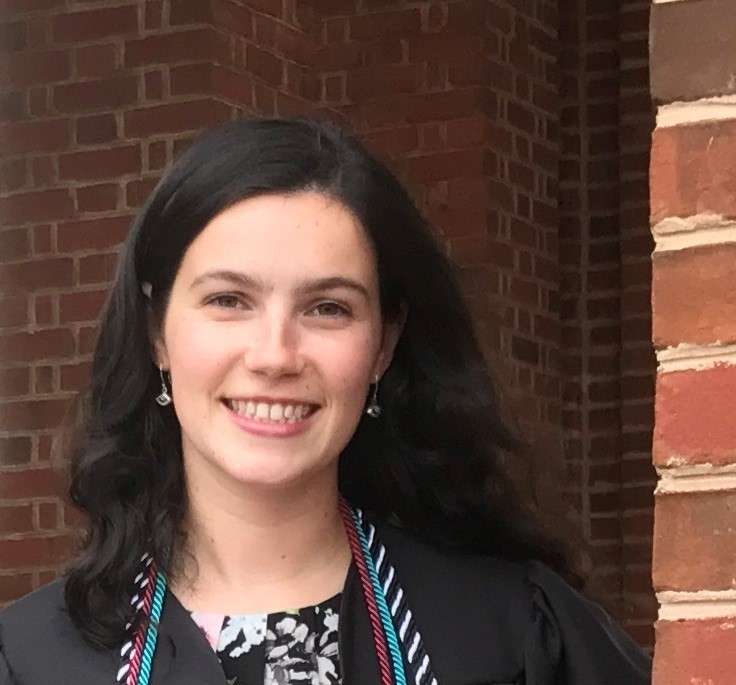}}]{Anna Goodridge} received the B.S. degree in Mechanical Engineering from Johns Hopkins University, Baltimore, MD, USA, in 2017 and is currently working for the Laboratory for Computational Sensing and Robotics as a research engineer. She has previously worked in research and development teams in the consumer product industry and her current work is with applications in surgical robotics and hardware design.
\end{IEEEbiography}

\vspace{-1cm}
\begin{IEEEbiography}[{\includegraphics[width=1in,height=1.25in,clip,keepaspectratio]{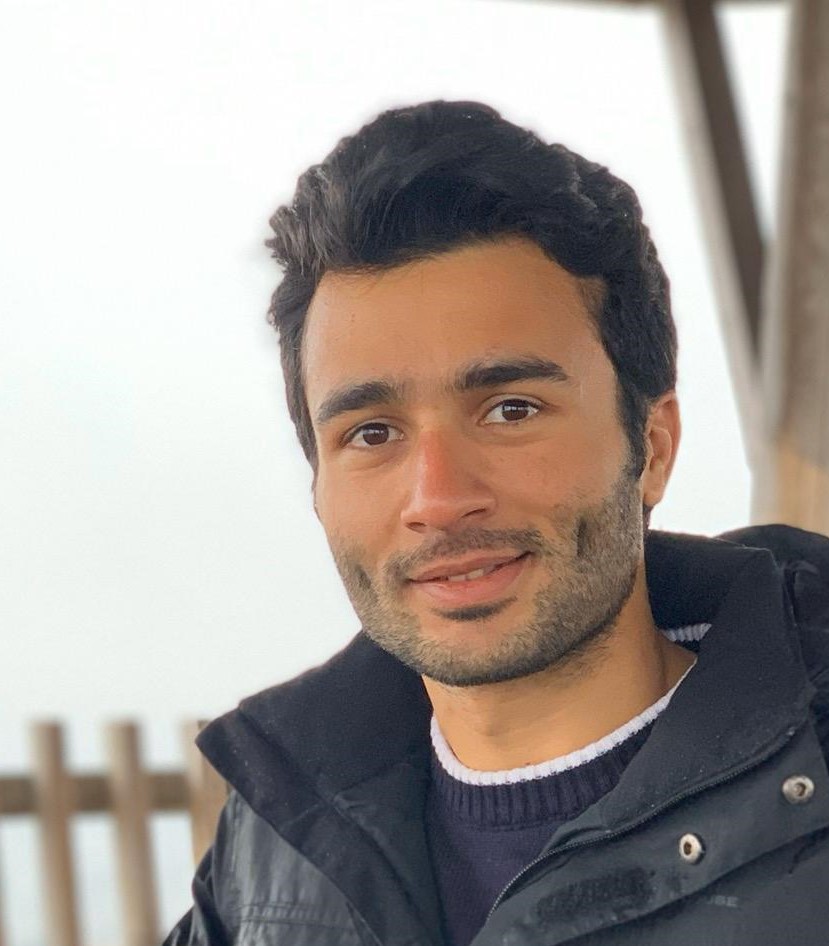}}]{Mojtaba Esfandiari} received the B.Sc. degree in Mechanical Engineering from Amirkabir University of Technology (Tehran Polytechnic), Tehran, Iran, in 2014, and the M.Sc. degree in Mechanical Engineering from Sharif University of Technology, Tehran, Iran, in 2017. He is currently pursuing the Ph.D. degree in Mechanical Engineering at Johns Hopkins University, Baltimore, MD, USA. He also worked as a research assistant at the Project neuroArm, the University of Calgary, Calgary, AB, Canada, during 2018-2021. 

His research interests include modeling, optimization, and control of surgical robotic systems. 
\end{IEEEbiography}

\vspace{-1cm}
\begin{IEEEbiography}[{\includegraphics[width=1in,height=1.25in,clip,keepaspectratio]{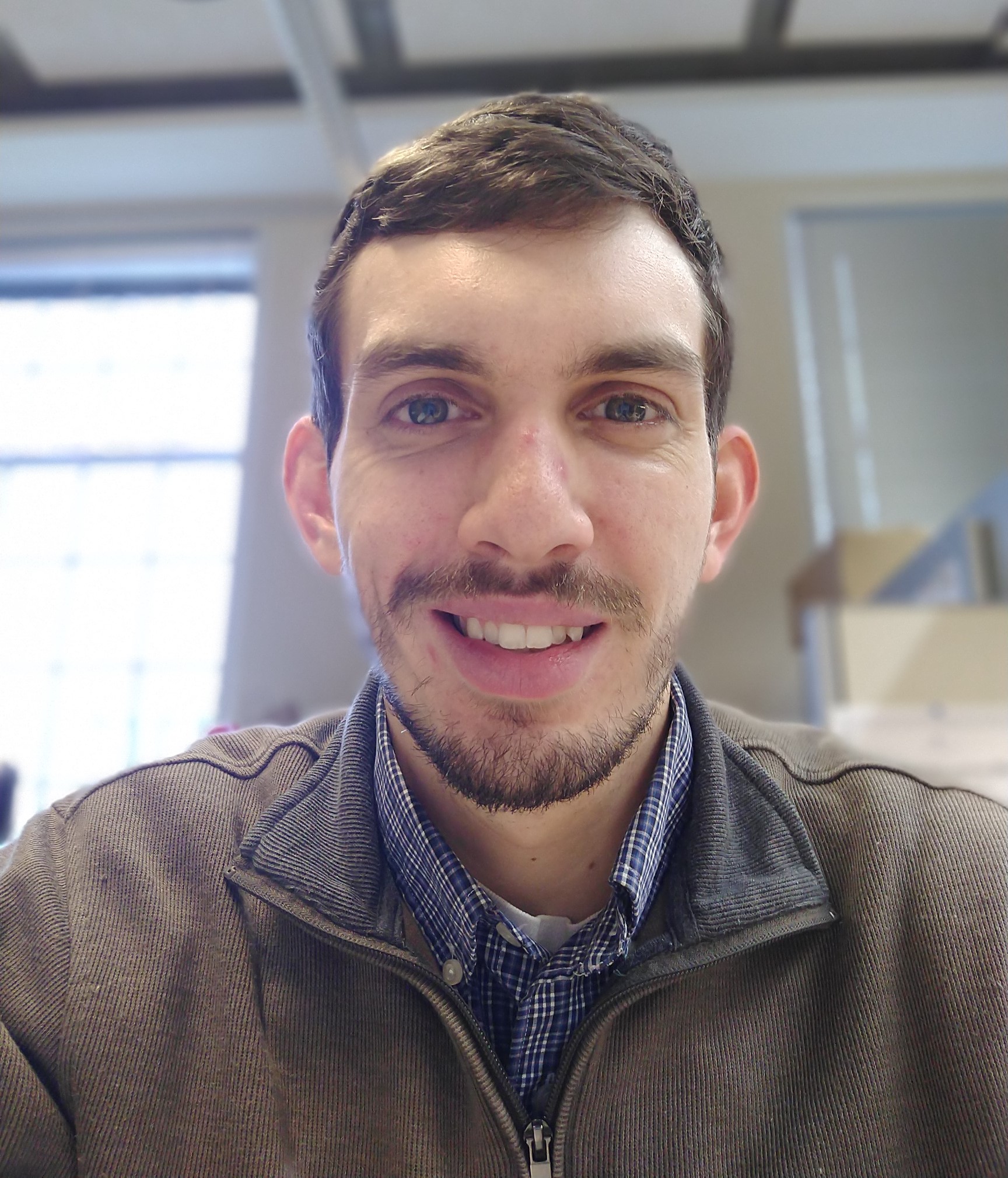}}]{Henry Phalen} received a B.S. degree in Bioengineering from the University of Pittsburgh, Pittsburgh, PA, USA in 2018 and a M.S.E. degree in Robotics from the Johns Hopkins University (JHU), Baltimore, MD, USA in 2021. He is currently a Ph.D. candidate in Mechanical Engineering at JHU.

His research interests include state estimation and control of dexterous robotic systems in constrained and safety-critical environments with a focus on demonstrating integrated systems for use in surgical procedures. 
\end{IEEEbiography}

\vspace{-1cm}
\begin{IEEEbiography}[{\includegraphics[width=1in,height=1.25in,clip,keepaspectratio]{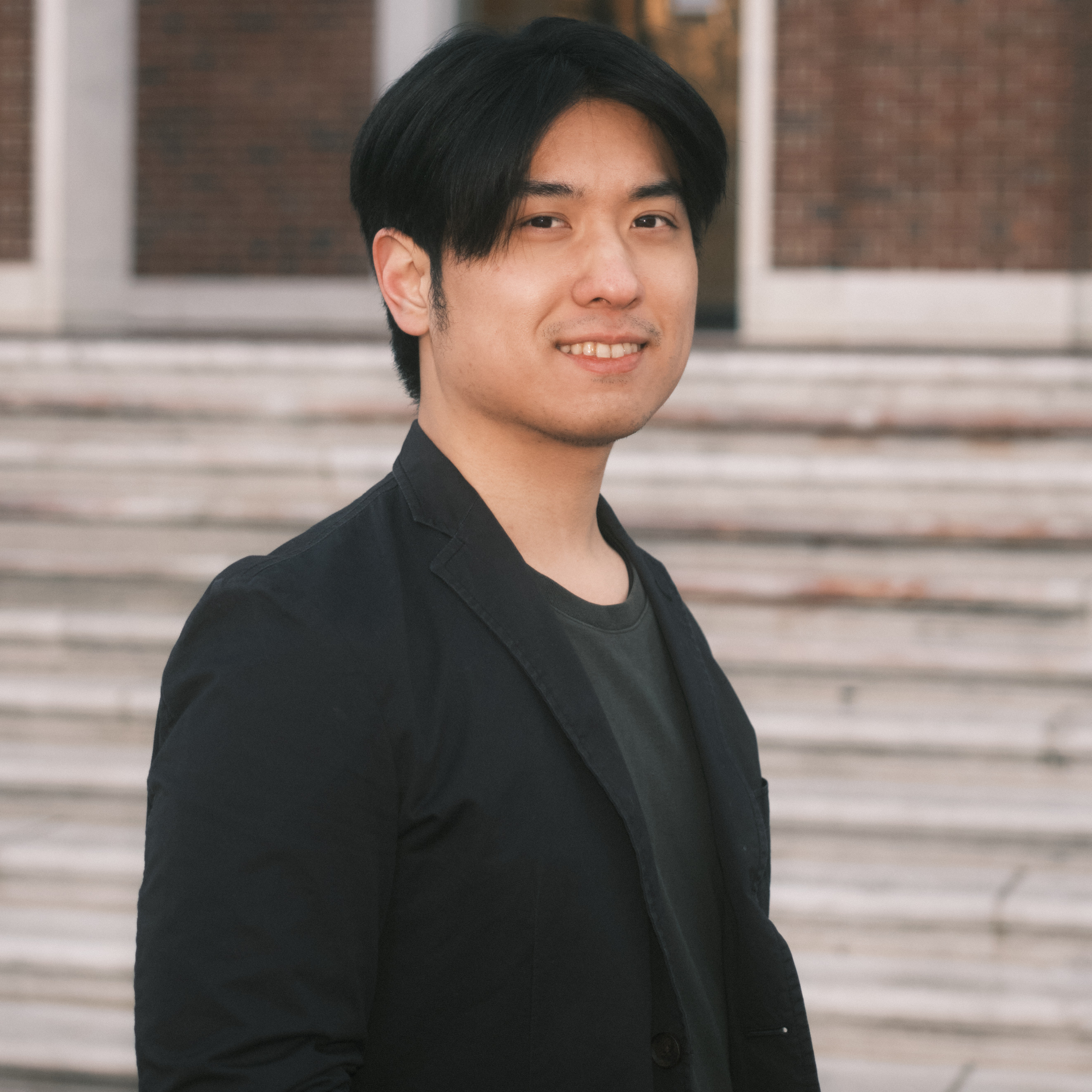}}]{Justin H. Ma} received the B.S degree in Mechanical Engineering from Georgia Institute of Technology, Atlanta, GA, USA, in 2019, and the M.S degree in Mechanical Engineering from Johns Hopkins University, Baltimore, MD, USA, in 2022. He is currently pursuing the Ph.D. degree in Mechanical Engineering at Johns Hopkins University, Baltimore, MD, USA. 

His research interests include design and control of continuum manipulators and mechanisms for minimally invasive surgical robotics in orthopaedics.  He has previously worked in the medical device industry developing implants for spinal fusion. 
\end{IEEEbiography}

\vspace{-1cm}
\begin{IEEEbiography}[{\includegraphics[width=1in,height=1.25in,clip,keepaspectratio]{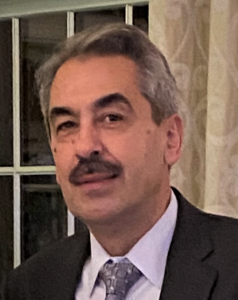}}]{Iulian Iordachita} (Senior Member, IEEE)  received
the M.Eng. degree in industrial robotics and the Ph.D.
degree in mechanical engineering from the University
of Craiova, Craiova, Romania, in 1989 and 1996,
respectively.

He is currently a Faculty Member with the Laboratory for Computational Sensing and Robotics,
Johns Hopkins University, Baltimore, MD, USA, and
the Director of the Advanced Medical Instrumentation and Robotics Research Laboratory. His current
research interests include medical robotics, image
guided surgery, robotics, smart surgical tools, and medical instrumentation.
\end{IEEEbiography}

\vspace{-12.6cm}
\begin{IEEEbiography}[{\includegraphics[width=1in,height=1.25in,clip,keepaspectratio]{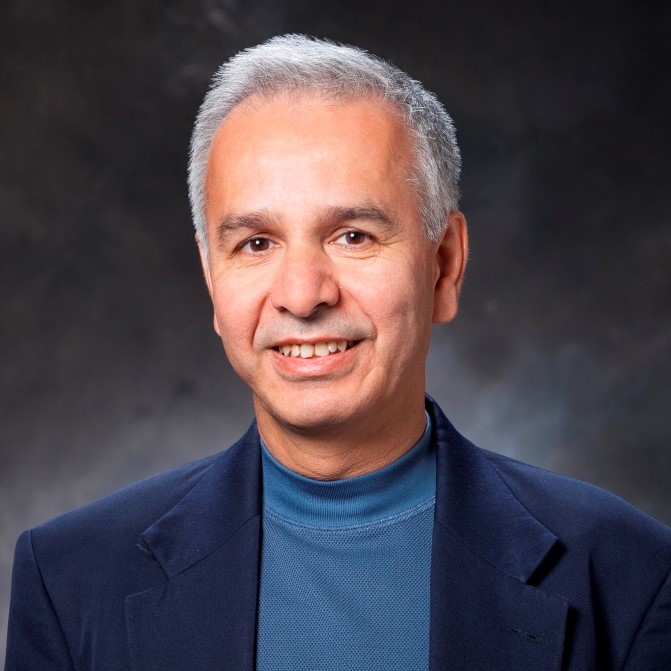}}]{Mehran Armand} (Member, IEEE)  received the
Ph.D. degree in mechanical engineering and kinesiology from the University of Waterloo, Waterloo, ON,
Canada, in 1998.

He is currently a Professor of Orthopaedic Surgery,
Mechanical Engineering, and Computer Science with
Johns Hopkins University (JHU), Baltimore, MD,
USA, and a Principal Scientist with the JHU Applied Physics Laboratory (JHU/APL). Prior to joining JHU/APL in 2000, he completed postdoctoral
fellowships with the JHU Orthopaedic Surgery and
Otolaryngology-Head and Neck Surgery. He currently directs the Laboratory
for Biomechanical- and Image-Guided Surgical Systems, JHU Whiting School
of Engineering. He also directs the AVICENNA Laboratory for advancing
surgical technologies, Johns Hopkins Bayview Medical Center. His laboratory
encompasses research in continuum manipulators, biomechanics, medical image analysis, and augmented reality for translation to clinical applications of
integrated surgical systems in the areas of orthopaedic, ENT, and craniofacial
reconstructive surgery.
\end{IEEEbiography}

\end{document}